\documentclass[twoside,11pt]{article}

\usepackage{jmlr2e}
\usepackage{graphicx}
\usepackage{epigraph}

\jmlrheading{1}{2000}{1-48}{4/00}{10/00}{Khyathi Raghavi Chandu}

\begin{document}

\title{Multilingual Multimodality: A Taxonomical Survey of Datasets, Techniques, Challenges and Opportunities}

\author{\name Khyathi Raghavi Chandu \email khyathi@meta.com \\
      \addr Meta \\
      \AND
      \name Alborz Geramifard \email alborzg@meta.com \\
      \addr Meta }


\editor{}

\maketitle

\begin{abstract}
Contextualizing language technologies beyond a single language kindled embracing multiple modalities and languages. Individually, each of these directions undoubtedly proliferated into several NLP tasks. Despite this momentum, most of the  multimodal research is primarily centered around English and multilingual research is primarily centered around contexts from text modality. Challenging this conventional setup, researchers studied the unification of multilingual and multimodal (MultiX) streams. The main goal of this work is to catalogue and characterize these works by charting out the categories of tasks, datasets and methods to address MultiX scenarios. To this end, we review the languages studied, gold or silver data with parallel annotations, and understand how these modalities and languages interact in modeling. 
We present an account of the modeling approaches along with their strengths and weaknesses to better understand what scenarios they can be used reliably. Following this, we present the high-level trends in the overall paradigm of the field. 
Finally, we conclude by presenting a road map of challenges and promising research directions. 
\end{abstract}


\section{Introduction}

\begin{quote}
\textit{Democratic reach of ubiquitous contexts from vision and language\textit{(s)} mitigates digital divide and alleviates cultural inclusion.}
\end{quote}

Our world contextualizes information in various modalities, the expression of which varies based on the medium i.e., languages. The recent revolutions of (i) aggrandizing NLP technologies to multiple languages \cite{DBLP:journals/csur/DabreCK20} to broaden the stakeholders to speakers of over 7000 living languages, and (ii) deriving context from multiple modalities \cite{DBLP:journals/pami/BaltrusaitisAM19} have mostly been independent of each other. However, to curtail the inequality of information access and cultural biases in building resources, discern multilingual and multimodal (MultiX) technologies is crucial. The status quo is still that English is the most prevalent language in grounded language agents. This creates a bottleneck for sharing information across models from other languages. 

\begin{figure}[t!]
\centering
\includegraphics[trim=4cm 8.5cm 2cm 1cm,clip,width=0.9\linewidth]{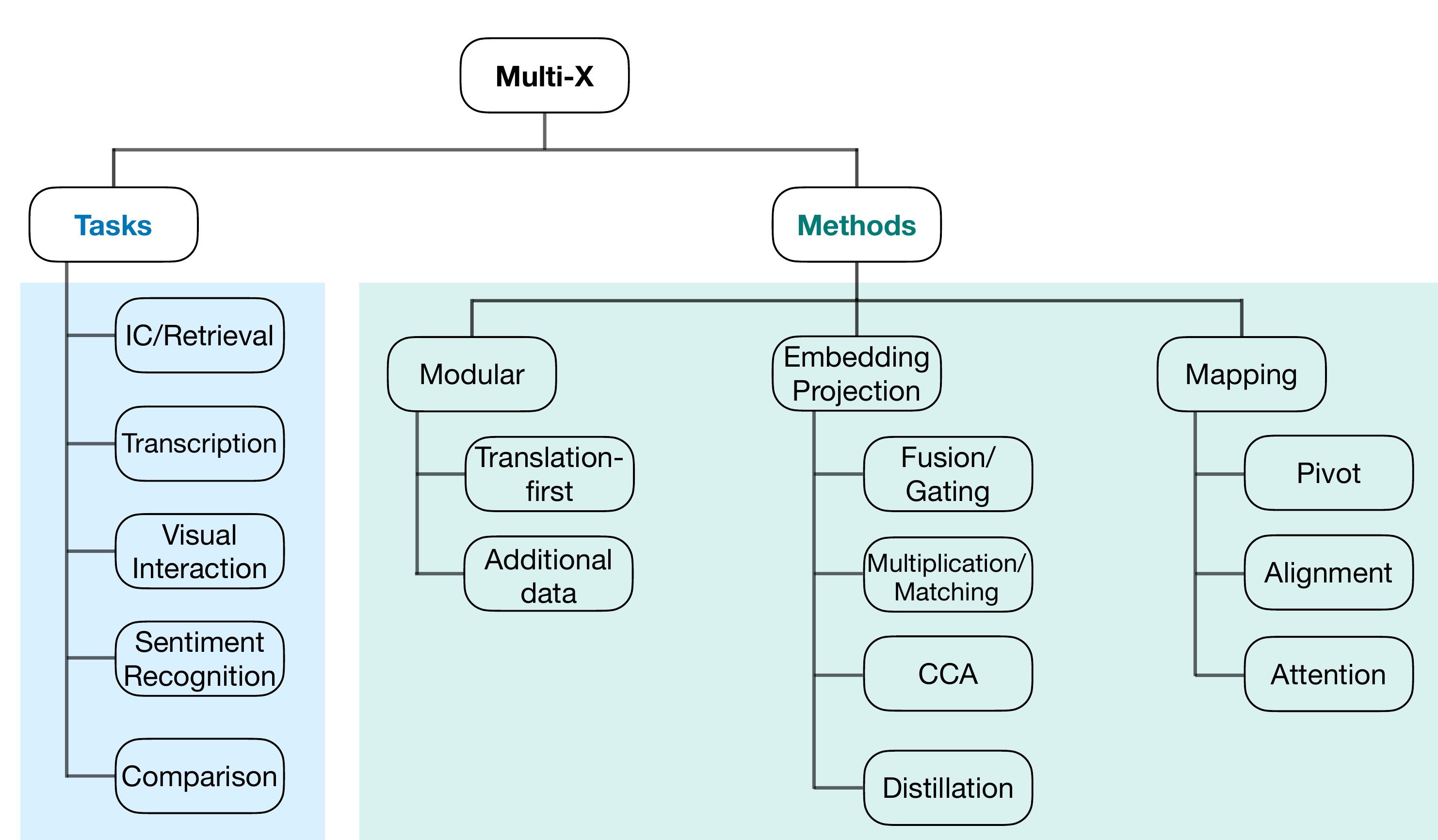}
\caption{Organization of the survey}
\label{fig:org}
\end{figure}

Despite the conventional trend of independent approaches to multilinguality and multimodality, researchers have sporadically yet consistently studied a confluence of these.
This paper presents a comprehensive survey of tasks, datasets and methods tackling MultiX scenarios. 
We aggregated papers from the last 60 years based on keyword search for variants of each of multiple modalities and languages in the titles and abstracts. Upon this, we filtered the papers manually that conduct research on both multilinguality and multimodality, and annotated them for MultiX specifics.
Distilling these learnings, we characterize an ontology of tasks and methods as depicted in Figure \ref{fig:org}. The confluence of both `X's brings about distinct challenges in dataset collection and methodologies which inspires this setup. Note that the names indicate how the modalities interact with several languages (as opposed to a inter-modal interactions with a single language). The rest of the survey is organized and presented according to the categories in Figure \ref{fig:org}.

The main goal of this paper is to present insights into what distinguishes MultiX in terms of resources and modeling.
Section \ref{sec:datasets} discusses the various tasks and datasets currently available. First, most MultiX datasets are built on top of existing visual or multimodal source datasets. To review this, we audit the source of the visuals used and the languages the corresponding tasks are explored in. Second, with the unique aspect of textual data present in multiple languages, we survey whether this data is \textit{parallel} (each instance is present across languages), and third, if they are \textit{gold} (human annotated), \textit{silver} (weakly annotated with off-the-shelf tools and models), or  \textit{direct} (scraped as is from sources). \S \ref{sec:methods}  categorizes the modeling approaches for MultiX into 9 classes based on interactions between languages and modalities. A distinctive feature of MultiX in modeling is leveraging annotations from one language while predicting in another language, which is referred to as cross-lingual. In contrast, some methods are developed and scaled to several languages independent of one another. Finally \S \ref{sec:future} presents existing challenges in the bleeding edge of research along with guided promising directions to improve MultiX.

\section{Tasks and Datasets}
\label{sec:datasets}

This section presents various tasks and the corresponding  datasets as shown in Table \ref{tab:datasets}.

\begin{table*}[tbh]
\centering
\resizebox{0.99\textwidth}{!}{%
\begin{tabular}{llllll}
\hline
\textbf{Task} & \textbf{Src} & \textbf{Citation} & \textbf{Languages} & \textbf{Gold/Silver} & \textbf{Parallel?} \\ \hline
Image Captioning, & IAPR & \cite{grubinger2006iapr} & de, es & Gold & Yes \\
Retreival & Pascal Sentences & \cite{DBLP:conf/emnlp/FunakiN15} & ja & Gold & Yes \\
 & Flickr8k & \cite{DBLP:conf/mir/LiLDL16} & zh & Gold Test & Yes \\
 & Flickr30k & \cite{DBLP:conf/mm/LanLD17} & zh & Gold Test & Yes \\
 &  & \cite{nakayama-etal-2020-visually} & ja & Gold & Yes \\
 &  & \cite{DBLP:conf/acl/ElliottFSS16} & de, fr, cs & Gold & Yes \\
 & Multi30k & \cite{DBLP:conf/lrec/LalaS18} & de, fr & Silver + Gold & No \\
 & MSCOCO & \cite{DBLP:journals/tmm/LiXWLJYX19} & zh & Gold test & Yes \\
 &  & \cite{DBLP:conf/acl/YoshikawaST17} & ja & Silver & Yes \\
 &  & \cite{DBLP:conf/naacl/RajendranKCR16} & de, fr & Silver & Yes \\
 &  & \cite{DBLP:conf/acl/HitschlerSR16} & de & Gold & Yes \\
 &  & \cite{DBLP:conf/wmt/ElliottFBBS17} & de, fr & Gold test & Yes \\
 & Visual Genome & \cite{parida-etal-2021-nlphuts} & hi & Gold & Yes \\
 & CC & \cite{caglayan-etal-2021-cross} & de & Silver & Yes \\
 & Flickr30k, MSCOCO, CC & \cite{DBLP:conf/cvpr/SurisEV22} & 52 langs & Silver & Yes \\
 \hline
Transcription & Video Interviews & \cite{jones-muftic-2020-endangered} & kqz, naq & Direct & No \\
 & TED Talks & \cite{karakanta-etal-2020-must} & de, es, fr, it,  &  &  \\
  &  &  & nl, pt, ro & Direct & No \\
 & Wilderness & \cite{DBLP:conf/icassp/Black19} & 700 languages & Direct & Yes (Silver) \\
 & Twitch.tv & \cite{fu-etal-2017-video} & zh-tw, en & Direct & No \\
 & YouTube & \cite{DBLP:journals/corr/abs-1811-00347} & pt, en & Direct & No \\
 &  Kinetics-600 & \cite{DBLP:conf/iccv/WangWCLWW19} & zh, en & Gold & Yes (Subset) \\
\hline
Description & IKEA website & \cite{zhou-etal-2018-visual} & en, de, fr & Direct & Close \\
 & Euronews & \cite{afli-etal-2017-multinews} & en, fr, ar, de, es,  &  &  \\
 &  &  & it, pt, tr, ua & Direct & Yes (Silver) \\
\hline
VQA & ImageNet+MSCOCO & \cite{ramnath-etal-2021-worldly} & hi, te & Silver & Yes \\
 &  & \cite{DBLP:conf/nips/GaoMZHWX15} & zh & Gold & No \\
 &  & \cite{koeva-2021-multilingual} & bg, hr, da, nl, en,  & &  \\
 &  &  & fi, fr, el, it, lt, &  &  \\
 &  &  & pl, pt, ro, sk, sl,  &  &  \\
 &  &  & es, sv, sq, is, he, sr & Silver & Yes \\
 & VQA 2.0 & \cite{raj-khan-etal-2021-towards-developing} & hi, bn, es, de, fr, & & \\
  & & & en-hi, en-bn, en-es,  &  &  \\
  & & & en-de, en-fr & Silver & Yes \\
 &  & \cite{gupta-etal-2020-unified} & hi, en-hi & Silver & Yes \\
 & Visual Genome & \cite{shimizu-etal-2018-visual} & ja & Gold & Yes (Subset) \\
 & & \cite{DBLP:conf/acl/PfeifferGKS0VG22} & en, de, bn, pt & Gold & Yes \\
 & & & ru, zh, ko, id & & \\
\hline
Dialog & Matterport3D & \cite{ku-etal-2020-room} & hi, te & Gold & No \\
 & Pentomino puzzle & \cite{zarriess-etal-2016-pentoref} & en, de & Gold & Yes \\
\hline
Comparison &  & \cite{liu-etal-2021-visually} & id, zh, sw, ta, tr & Gold & No \\
\hline
Misc &  & \cite{alberts-etal-2021-visualsem} & ar, zh, nl, en,  &  &  \\
 &  &  & fa, fr, de, it,  &  &  \\
 &  &  & ko, pl, pt, ru, &  &  \\
 &  &  & es, sv & Silver & Yes \\
 &  & \cite{moneglia-etal-2014-imagact} & it, zh, es & Silver + Gold & No \\
 &  & \cite{gella-etal-2019-cross} & de, es & Silver + Gold & Yes \\
Sentiment Recogn. & Youtube monologues & \cite{bagher-zadeh-etal-2020-cmu} & es, pt, de, fr & Gold & No \\
\hline
\end{tabular}
}
\caption{\small Datasets along with languages (ISO codes) spanned, gold/silver nature of annotations and parallism across languages.}
\label{tab:datasets}
\end{table*}


\subsection{Image Captions: }

\paragraph{IPAR-based: }
\citet{grubinger2006iapr} collect a dataset of 
images of various locations and actions accompanied with captions in three languages including English, German, and Spanish.

\paragraph{Flickr-based: }
Instead of semi-synthetic captions, 
Flickr30k Entities \cite{plummer2015flickr30k} is extended to Japanese (F30kEnt-JP) \cite{nakayama2020visually} with phrase-to-region linking so the cross-lingual phrase-to-phrase relations can be exploited meaningfully. 
They are also extended to Chinese (Flickr8k-CN \cite{DBLP:conf/mir/LiLDL16}, Flickr30k-CN \cite{burger-etal-2003-communicative}) with a semi-automatic and human post-editing step, and into Multi30k in German, French, Czech \cite{DBLP:conf/acl/ElliottFSS16} with human translations. A similar adaptation from images and English captions of Flick8k dataset is extended to automatically create Chinese captions \cite{DBLP:conf/mir/LiLDL16} and a semi-human created test set.

\paragraph{MS-COCO-based: }
The MS-COCO captions \cite{DBLP:conf/eccv/LinMBHPRDZ14} are extended to Chinese \cite{DBLP:journals/tmm/LiXWLJYX19}, Japanese \cite{DBLP:conf/acl/YoshikawaST17}, German \cite{DBLP:conf/naacl/RajendranKCR16, DBLP:conf/acl/HitschlerSR16} and French \cite{DBLP:conf/naacl/RajendranKCR16}. Excepting  \citet{DBLP:conf/acl/HitschlerSR16}, the rest are silver datasets, i.e., translated automatically from English captions. 

\paragraph{Visual Genome-based: }
The annotations for Visual Genome \cite{DBLP:journals/ijcv/KrishnaZGJHKCKL17} are partly reused by automatically translating them to Hindi using segment NMT model \cite{parida2018translating}.They are automatically translated to Hindi using segment NMT model \cite{parida2018translating} and then humans edit the translation in based on the image. 
The challenge test set includes ambiguous English words (ambiguity  determined by embedding similarity), which can be resolved by visual context alone. The same is replicated for Malayalam Visual Genome \footnote{\url{https://ufal.mff.cuni.cz/malayalam-visual-genome/wat2021-english-malayalam-multi}}.


\paragraph{Retrieval: } 

\noindent The OpenCLIR Challenge \footnote{\url{https://www.nist.gov/itl/iad/mig/openclir-challenge}} is aimed to locate snippets of text and speech in documents of low-resourced languages such as Swahili. 
Several of the aforementioned captioning datasets are also studied for caption-retrieval in tandem to generation.

\subsection{Transcriptions: }

Inspired from speech translation dataset MuST-C \cite{DBLP:conf/naacl/GangiCBNT19} with sentence-level transcriptions, \citet{karakanta-etal-2020-must} released MuST-Cinema for subtitle generation of audio-visual content. It includes audio, transcription, translation triplets of TED talks in 7 languages including German, Spanish, French, Italian, Dutch, Portugese, Romanian. Scaling this up with videos from the wild from YouTube, \citet{DBLP:journals/corr/abs-1811-00347} presented a large-sclae dataset of How2 with of instructional videos on topics with $\sim$ 8k clips and $\sim$ 2k hours. In addition to being multimodal, translations into Portugese are also collected making the dataset multilingual. 
These audio-visual recordings are valuable for (a) scaling multimodal multilingual resources and (b) maximizing the utility of available data in low resourced languages.
(a) For scaling, \citet{jones-muftic-2020-endangered} collected interview recordings from audio visual legacy media in N|uu, Kora and Khoekhoe languages, emphasizing on the ethical steps to gather and publish such a dataset.
(b) For utilizing available data, \citet{DBLP:conf/icassp/Black19} massively scales to 700 languages with around 20 hours of speech for each language based on religious recordings from YouTube   annotated with a multi-pass alignment technique. In addition to the video transcriptions, the chats based on videos signal important events or highlights. \citet{fu-etal-2017-video} present a dataset with 300 videos of 30-50 minute games from Twitch.tv channels based on League of Legends championships and their corresponding $\sim$~7k chats to study video highlight prediction. \citet{DBLP:conf/iccv/WangWCLWW19} present VATEX based on the Kinetics-600 actions dataset \cite{DBLP:journals/corr/KayCSZHVVGBNSZ17} to perform the tasks of video captioning and video-guided machine translation.

\paragraph{Online descriptions: }
Internationally available online shopping platforms are a rich source of product descriptions in multiple languages ranging beyond simple image captions describing product semantics and their usage.
\citet{zhou-etal-2018-visual} present a parallel dataset of product descriptions from IKEA’s and UNIQLO’s websites in 3 languages including English, French, and German. 
In contrast to image captioning, the descriptions are not exactly parallel and are longer than captions.
Similarly, news outlets provide internationally viable descriptions across multiple languages. \citet{afli-etal-2017-multinews} introduced MultiNews dataset comprising of images and their descriptions spanning 9 European languages sourced from Euronews website.
The articles are not automatic translations but are aligned finely over sentences from human written texts in different languages.



\subsection{Visual Interaction: }

\paragraph{Visual Question Answering: }

\citet{raj-khan-etal-2021-towards-developing} extended VQA 2.0 \cite{DBLP:conf/cvpr/GoyalKSBP17} to a silver dataset with 6  languages using Google translation and 5 pairs of code-switched languages using matrix language frame theory \cite{myers1997duelling}. 
Prior to this, \citet{gupta-etal-2020-unified} translated these questions to Hinglish using question generation method by  \citet{DBLP:conf/conll/GuptaLEB18}. 
In addition to text, \citet{ramnath-etal-2021-worldly} created a synthetic dataset for spoken-VQA in 3 languages including English, Hindi, and Turkish.
To increase question diversity, \citet{DBLP:conf/nips/GaoMZHWX15} crowdsourced FM-IQA dataset based on COCO images with unconstrained  human annotations for questions and answers. 
Later, \citet{shimizu-etal-2018-visual} crowdsourced a Japanese visual question answering dataset with $\sim$700k QA pairs based on Visual Genome and Japanese question types, which are matched to the  English versions to model cross-lingual transfer.

\paragraph{Vision-Language Navigation: }
Extending Room2Room \cite{DBLP:conf/cvpr/AndersonWTB0S0G18}, inspired from localized narratives. 
\citet{ku-etal-2020-room} introduce a new dataset RoomXRoom in Hindi and Telugu. \citet{zarriess-etal-2016-pentoref} also propose PentoRef with around 20k utterances in English and German transcribed and annotated with referring expressions for reference resolution and generation.

\subsection{Sentiment/Emotion Recognition: }
\citet{bagher-zadeh-etal-2020-cmu} introduced CMU-MOSEAS based on Youtube monologues with $\sim$40k sentences in 4 languages including annotations for around 20 labels for sentiment, emotion, attributes etc.,

\subsection{Comparison: }
There is a stream of research on pivoting over the visual representations to map multiple languages (discussed in \S \ref{sec:methods}). While this hypothesis is justifiable for a majority of concepts, it does not cover culturally relevant and distinguishable concepts. 
Comparing similar yet distinct words based on cultural relevance and ambiguity is a unique challenge to MultiX. To address this, 
\citet{liu-etal-2021-visually} introduce MARVL, focusing on culturally relevant concepts in the corresponding nativities for 6 languages of Indonesian, Mandarin Chinese, Swahili, Tamil, and Turkish.
\citet{DBLP:conf/wmt/ElliottFBBS17} introduce Ambiguous COCO with a test set of ambiguous words in German, French and English.
and \citet{gella-etal-2019-cross} present the MultiSense dataset for disambiguating verbs in different languages in German and Spanish.

\paragraph{Videos: }
So far we have seen several datasets for static visuals, which can be extended to videos \cite{huang-etal-2021-multilingual}. This Multi-HowTo100M videos dataset includes 1.2 M videos, their corresponding  speech and transcriptions from 9 languages which can be used for massive pretraining. 

\subsection{Miscellaneous Datasets}

This section covers more of the datasets in MultiX that are not covered categorically in \S \ref{sec:datasets}. 
IMAGACT \cite{moneglia-etal-2014-imagact} is based on an ontology of actions from spontaneous speech over visuals. It is also intended to be used as a multilingual dictionary of images. 
\citet{gupta-etal-2021-vita} worked towards building a language-ontology based sets of images for object detection and segmentation. Dominant classes and thematic domains are extracted from the ontology and they are used to retrieve similar images from the web. The effort targeted 20 European languages.
\citet{alberts-etal-2021-visualsem} created a knowledge graph over 900k unique images with 1.3M multilingual gloss for 14 languages.

\section{Techniques}
\label{sec:methods}


\begin{table*}[tbh]
\centering
\resizebox{0.99\textwidth}{!}{%
\begin{tabular}{llllll}
\hline
\textbf{Modeling} &  &  &  &  &  \\
\textbf{Category} & \textbf{Sub-category} & \textbf{Work} & \textbf{Cross-lingual?} & \textbf{AV Features} & \textbf{Text Features} \\
\hline
Modular & Translation-first & \cite{arora-etal-2020-investigative} & No & MFCC (speech) & Phrase \\
 &  & \cite{route-etal-2019-multimodal} & No & MFCC (speech) & LSTM \\
 &  & \cite{ramnath-etal-2021-worldly} & Yes & MFCC (speech) & LSTM \\
 &  & \cite{de-melo-weikum-2010-providing} & Yes & - & Surface forms \\
 & Additional data & \cite{gupta-etal-2021-vita} & Yes & Faster R-CNN & mBART \\
 &  & \cite{lala-etal-2018-sheffield} & Yes & ResNet-50 & MUSE \\
 &  & \cite{calixto-liu-2017-incorporating} & Yes & VGG-19 & RNN \\
 \hline
Embedding & Fusion & \cite{DBLP:conf/iclr/0001C0USLZ20} & No & ResNet-50 & Transformer \\
Projection &  & \cite{parida-etal-2021-nlphuts} & Yes & CNN & LSTM \\
 &  & \cite{caglayan-etal-2021-cross} & Yes & Faster R-CNN & Transformer \\
 &  & \cite{parida-etal-2020-odianlps} & Yes & InceptionResNetv2 & Transformer \\
 &  & \cite{gella-etal-2019-cross} & Yes & ResNet-34 & Word2Vec \\
 &  & \cite{fu-etal-2017-video} & No & ResNet34-LSTM & Char-LSTM \\
 & Multiplication & \cite{susanto-etal-2021-rakutens} & Yes & UVR & mBART \\
 &  & \cite{fei-etal-2021-cross} & No & CLIP image encoder & CLIP text encoder(English)+SBERT \\
 &  &  &  & & paraphrase-multilingualmpnet-base-v2 \\
 &  &  &  & & (for other languages) \\
 &  & \cite{ku-etal-2020-room} & Yes & EfficientNetB4 CNN & mBERT \\
 &  & \cite{shi-etal-2019-visually} & No & ResNet-101 & FastText \\
 &  & \cite{mohammadshahi-etal-2019-aligning-multilingual} & Yes & ResNet-152 & FastText \\
 & CCA & \cite{nakayama-etal-2020-visually} & Yes & Faster-RCNN, VGG16, & MeCab tokenizer, \\
 &  & & & ResNet50 & word vectors \\
 &  & \cite{rotman-etal-2018-bridging} & Yes & VGG-19 & Skip-gram \\
 & Distillation & \cite{raj-khan-etal-2021-towards-developing} & Yes & LXMERT & mBERT \\
\hline
Mapping & Pivot on image & \cite{huang-etal-2021-multilingual} & Yes & R(2+1) spatio-temporal & mBERT \\
 &  &  &  & CNN &  \\
 &  & \cite{huang-etal-2020-unsupervised-multimodal} & Yes & Faster-RCNN & XLM \\
 &  & \cite{kadar-etal-2018-lessons} & Yes & ResNet50 & GRU \\
 &  & \cite{gella-etal-2017-image} & Yes & VGG-19 & GRU \\
 &  & \cite{calixto-liu-2017-sentence} & Yes & VGG-19 & GRU \\
 & Alignment & \cite{fei-etal-2021-cross} & Yes & VL-BERT & VL-BERT \\
 & & \cite{DBLP:conf/cvpr/SurisEV22} & Yes & ImageNet & BPE \\
 &  & \cite{nishihara-etal-2020-supervised} & Yes & ResNet-50 & Kytea for segmentation \\
 &  & \cite{huang-etal-2019-multi} & Yes & Faster-RCNN & mBERT \\
 & Attention & \cite{ramnath-etal-2021-worldly} & No & Faster R-CNN & IaK \\
 &  & \cite{singh-etal-2021-multiple} & Yes & CNN & RNN \\
 &  & \cite{mitzalis-etal-2021-bertgen} & Yes & VL-BERT & mBERT \\
 &  & \cite{imankulova-etal-2020-towards} & Yes & ResNet-50 & RNN \\
 &  & \cite{bagher-zadeh-etal-2020-cmu} & No & Conv1D + Transformer & Conv1D + Transformer \\
 &  & \cite{gupta-etal-2020-unified} & Yes & R-CNN & fastText, CNN \\
 &  & \cite{zhou-etal-2018-visual} & Yes & ResNet & LSTM \\
 &  & \cite{shimizu-etal-2018-visual} & Yes & VGG-19 & LSTM \\
\hline
Other &  & \cite{karakanta-etal-2021-simultaneous} & No & CNN & CNN \\
 &  & \cite{liu-etal-2021-visually} & No & baselines & baselines \\
 &  & \cite{karakanta-etal-2020-must} & Yes & No images just NMT & Transformer \\
 &  & \cite{vilares-etal-2020-bringing} & No & X, Y coordinates & (linguistic ftrs) \\
 \hline
\end{tabular}
}
\caption{\small Categories of modeling approaches serving cross-lingually with feature representations for modalities and languages.}
\label{tab:methods}
\end{table*}


This section presents a hierarchical categorization of modeling approaches as depicted in Table \ref{tab:methods}.

\subsection{Modular: }

\paragraph{Translation-first: } \mbox{} \\

\begin{figure}[tbh]
\centering
\includegraphics[trim=0cm 0cm 0cm 0cm,clip,width=0.8\linewidth]{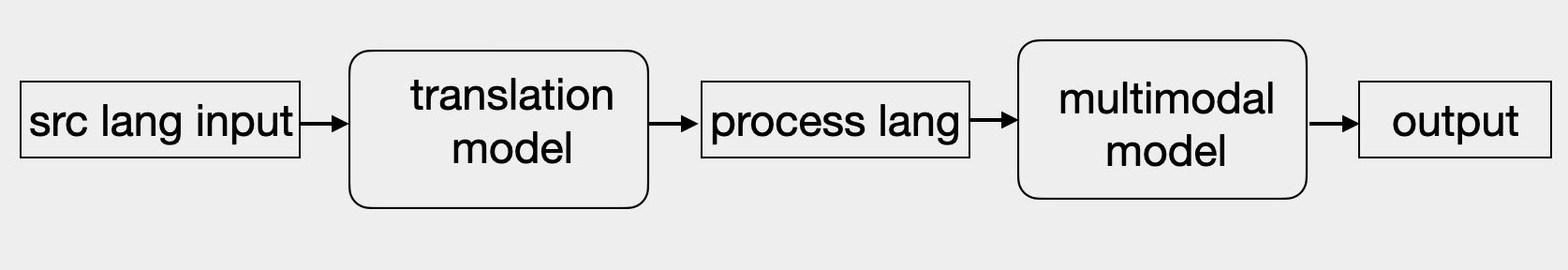}
\caption{Translation-first approaches}
\label{fig:translation}
\end{figure}

A broad idea of the modular approach to the translation-first approaches is presented in Figure \ref{fig:translation}. As we can see the first step here is to translate the source language input (often times unimodally in textual modality) into the processing language which is equipped with the multimodal model to predict the output. This chain of models can either be independently trained or jointly trained end to end. Often times, the strategy is to translate into the language that has a stronger performing multimodal model in a high-resource language, which resolves to be English most of the times.

Most methods for spoken-VQA \cite{DBLP:journals/corr/ZhangDTMG17} includes a two-step process of performing ASR and then answer prediction. In contrast, \citet{ramnath-etal-2021-worldly} proposes to use speech embeddings directly to predict the answer. This is particularly useful when there are limited resources to perform ASR and translation from low-resourced languages. Similarly, \citet{huang-etal-2020-unsupervised-multimodal} use Google translate to translate the COCO captions to French and German for pretraining followed by visual pivoting for training. In the same spirit of modularity, \citet{arora-etal-2020-investigative} approach this with individual components for speech processing, translating the query and then performing retrieval in the target language. Dictionaries (from the Kamusi project \url{https://kamusi.org/)}) and phrase based SMT  \cite{koehn2003statistical} models with the MOSES toolkit \cite{DBLP:conf/acl/KoehnHBCFBCSMZDBCH07} is used for query translation followed by keyword search \cite{DBLP:conf/interspeech/TrmalWPZGWMXPK17} for retrieval.  \citet{de-melo-weikum-2010-providing} also translate Wikipedia pages first using 
Google Translate to respond to a user’s lexical queries by responding with multilingual and multimodal information using token similarity (for the first sense) and etymological approximation.
\citet{route-etal-2019-multimodal} use OpenNMT seq2seq framework to multitask decoder with input as text and MFCC features for the task of TTS and use the hidden state of the first output to  predict IPA.
\citet{DBLP:conf/icml/Bugliarello0PRE22} introduce the IGLUE benchmark by aggregating pre-existing datasets and also collecting new ones across several tasks such as —visual question answering,
cross-modal retrieval, grounded reasoning, multimodal entailment etc.,. They observe that `` translate-test transfer is superior to zero-shot transfer and that few-shot learning is hard to harness for many tasks" making this approach both a very strong baseline and state of the art.

\paragraph{Additional data (weak labels): } \mbox{} \\

\begin{figure}[!t]
\centering
\includegraphics[trim=0cm 0cm 0cm 0cm,clip,width=0.8\linewidth]{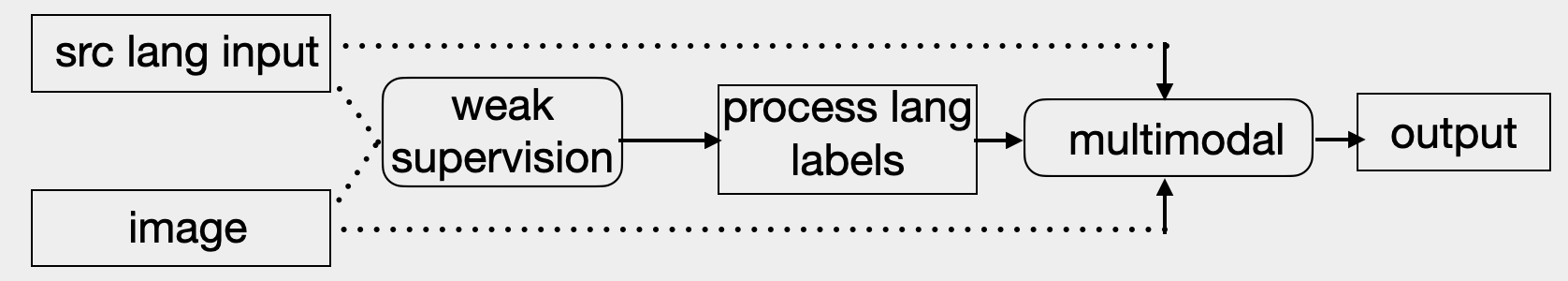}
\caption{Weak supervision based approaches}
\label{fig:weak}
\end{figure}

Annotating the data with additional labels that can assist in modeling is another way to approach multiple modalities and languages. This approach on a broad level is presented in Figure \ref{fig:weak}. In the case of vision and language tasks, either the images are annotated with labels in a processing language or the source language input is annotated with additional labels in a processing language. These annotations are often obtained using off-the-shelf tools as weakly supervised labels. 
Upon the completion of this module, these additional annotations obtained are used along with the initial inputs are given to a multimodal model for the final predictions.

\citet{gupta-etal-2021-vita} use image recognition models to extract object tags in English in the first stage to subsequently use them as  weak labels in the encoder to decode in a different language. This cross-lingual learning from English tags helps map textual co-occurrences for different images in text modality in addition to the image information. 
\citet{calixto-liu-2017-incorporating} use global image features extracted from VGG19 as additional data to initialize the encoder and the decoder hidden states in the RNNs. 
\citet{barrault-etal-2018-findings} assimilate findings from various systems for multimodal machine translation and observe that data augmentation significantly improves the task performance. Along the same lines, \citet{lala-etal-2018-sheffield} augment n-best lists of translated data from French, German and Czech to English to train an NMT model. Despite being noisy, this augmentation of silver data  performs better than the corresponding baseline.

\subsection{Embedding Projections: }

\paragraph{Fusion and Gating: } \mbox{} \\

\begin{figure}[tbh]
\centering
\includegraphics[trim=0cm 0cm 0cm 0cm,clip,width=0.8\linewidth]{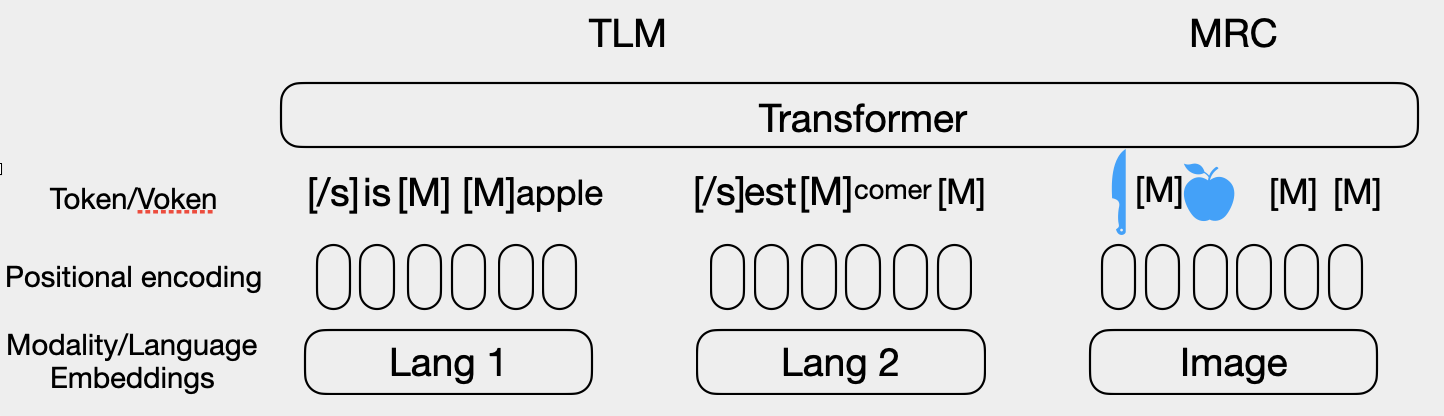}
\caption{Fusion based approaches}
\label{fig:fusion}
\end{figure}

Fusion based approaches combine the information from different modalities, and the text from different languages in a defined design. They can be categorized into early fusion in which the inputs are combined at the feature embedding level or late fusion where the projections are learnt by a deep task dependent network, which are combined in at a later position. While early fusion is considered an integration at the feature level, the late fusion is considered an integration at the semantic level. 
An example of the approaches based on fusion is presented in Figure \ref{fig:fusion} in the recent era of pretraining. The signals from both the languages are integrated using a translation based objective described in detail in the following paragraph. The objective across modalities is optimized with image region based objectives.

\citet{DBLP:conf/iclr/0001C0USLZ20} introduce universal visual representations (UVR) to perform multimodal machine translation by leveraging a group of images that have similar topics contained in the source sentence. This similarity is determined by tf-idf scores followed by the fusion of the visual information with gating to predict the target translations. Similarly, \citet{parida-etal-2021-nlphuts} performs region specific captioning by fusing image features with region coordinates features. 
\citet{caglayan-etal-2021-cross} extend the TLM (translation language model) objective \cite{DBLP:conf/nips/ConneauL19} to include regional image features via VTLM model that concatenates the translations of both languages as input to the same encoder optimized for masked textual and visual tokens. 
The model is optimized for TLM and MRC (masked region classification) where the visual and textual tokens in the input are masked. 
In addition, a simple concatenation of image features and source language features also demonstrate decent performance and 
are also used as baselines for several tasks. For instance, \citet{parida-etal-2020-odianlps} represent images using InceptionResNetv2 and text using transformers and combine their representations to perform multimodal machine translation. Similarly, \citet{gella-etal-2019-cross} concatenate the visual and textual features for predicting the word sense in another language. Instead of direct concatenation, \cite{fu-etal-2017-video} add an MLP layer on the modality representations to predict highlights in a video. The recent era of pretraining also relies on these heavily silver standard data by using automatic translation tools to first translate the data \cite{DBLP:conf/cvpr/ZhouZW0LYL21} and use them for MultiX modeling.

\paragraph{Matching: } \mbox{} \\

\begin{figure}[tbh]
\centering
\includegraphics[trim=0cm 0cm 0cm 0cm,clip,width=0.6\linewidth]{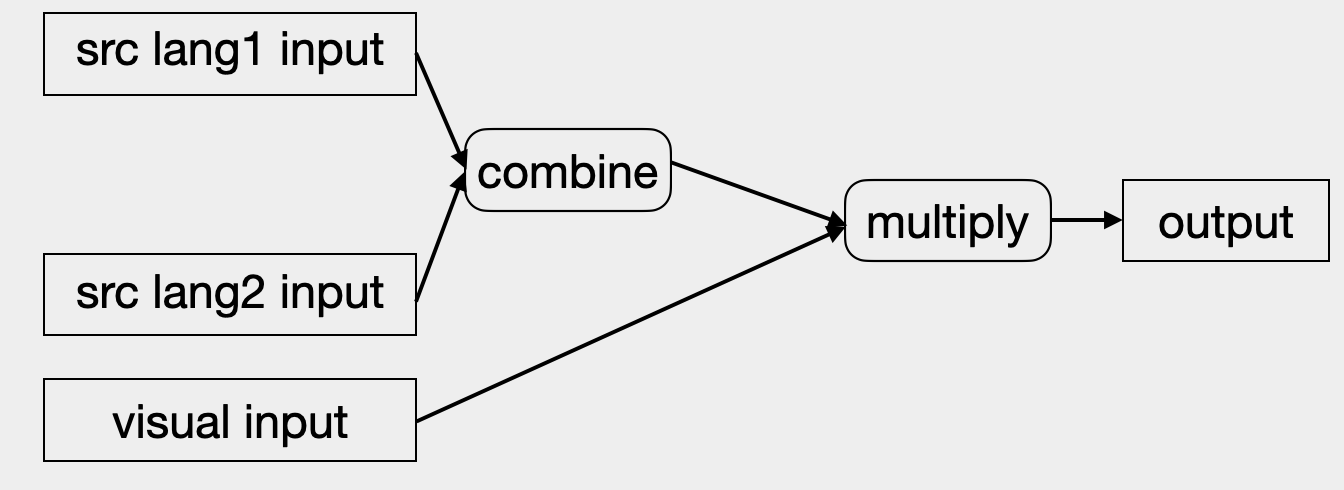}
\caption{Matching based approaches}
\label{fig:matching}
\end{figure}

The matching based techniques are also specific to the task at hand. The embedding projections for both the languages are first combined. This combine multilingual embedding is matched with the visual input to predict the output. An overview of these approaches is presented in Figure \ref{fig:matching}.

\citet{susanto-etal-2021-rakutens} use the universal visual representations described earlier and compute similarities with the language encodings obtained from mBART. 
Similarly, \citet{fei-etal-2021-cross} perform entity, sentence and image retrieval on VisualSem knowledge graph by computing the similarity between image embeddings
and text embeddings in multiple languages. The visual representation is derived using the CLIP image encoder \cite{DBLP:conf/icml/RadfordKHRGASAM21} and the textual representation is derived using the CLIP text encoder for English and SBERT
model paraphrase-multilingualmpnet-base-v2 \cite{DBLP:conf/emnlp/ReimersG19} for other languages. 
An agent similar to Reinforced Cross-Modal Matching \cite{DBLP:conf/cvpr/WangHcGSWWZ19} is adapted by replacing LSTMs with successive 1D convolutions to encode longer utterances for  multilingual navigation in RXR \cite{ku-etal-2020-room}. They note that despite 3 times the data, training a single multilingual agent on several languages  performs worse compared to their monolingual counterparts, while the multilingual agent outperforms with a  multitask setting combining annotations from Room2Room. 
For matching textual representations to visual inputs and combining constituents, \citet{shi-etal-2019-visually} optimize for the hinge triplet loss jointly by building constituency structures recursively \cite{DBLP:conf/acl/KleinK18} with a bottom-up score-sample-combine approach.  
The visual semantic embedding space \cite{DBLP:journals/corr/KirosSZ14} is created with cosine similarity based matching score in the joint space. 
Likewise, \citet{mohammadshahi-etal-2019-aligning-multilingual}  computes this similarity where the text representation is a combination of multiple languages with internal alignment using kNN based algorithm.
\citet{afli-etal-2017-multinews} use word based and named entity based scoring strategies to align news corpora from Euronews spread across 9 languages. 

\paragraph{Canonical Correlations: } \mbox{} \\

\begin{figure}[tbh]
\centering
\includegraphics[trim=0cm 0cm 0cm 0cm,clip,width=0.5\linewidth]{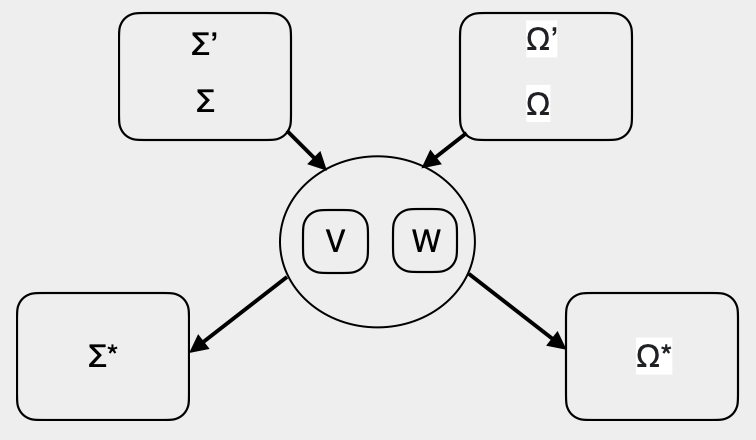}
\caption{CCA based approaches}
\label{fig:cca}
\end{figure}

Canonical Correlation Analysis (CCA) is a method to infer information from cross-covariance matrices. With multiple vectors of random variables along with correlations among these variables, then this method is used to find a linear combination of these variables with maximum correlation among themselves. This is depicted in Figure \ref{fig:cca}.

\citet{nakayama-etal-2020-visually} use a generalized CCA (GCCA)  \cite{DBLP:journals/ijcv/GongKIL14} to include English translation as an additional modality to perform phrase localization in images using Japanese text. This technique is earlier developed to use higher level semantics as the third modality to perform text-to-image alignment. Similarly, Japanese, English and images are the three modalities where the nearest embedding in the canonical subspace is retrieved. GCCA can thus perform cross-lingual retrieval.
Instead of including an additional modality, \citet{rotman-etal-2018-bridging} present partial CCA (PCCA) to maximize the canonical correlation of the multilingual descriptions in two languages conditioned on the shared variable of the image representation. 
The difference between PCCA and GCCA is that the former attempts to maximize the canonical correlations of all pairs of views whereas the latter condition two variables on the third shared one.
\citet{leviant2015separated} extended SimLex-999 and wordsim353 annotations to Italian, German, and Russian, which are later experimented for CCA performance by \citet{rotman-etal-2018-bridging}. 

\paragraph{Distillation: } \mbox{} \\

\begin{figure}[tbh]
\centering
\includegraphics[trim=0cm 0cm 0cm 0cm,clip,width=0.65\linewidth]{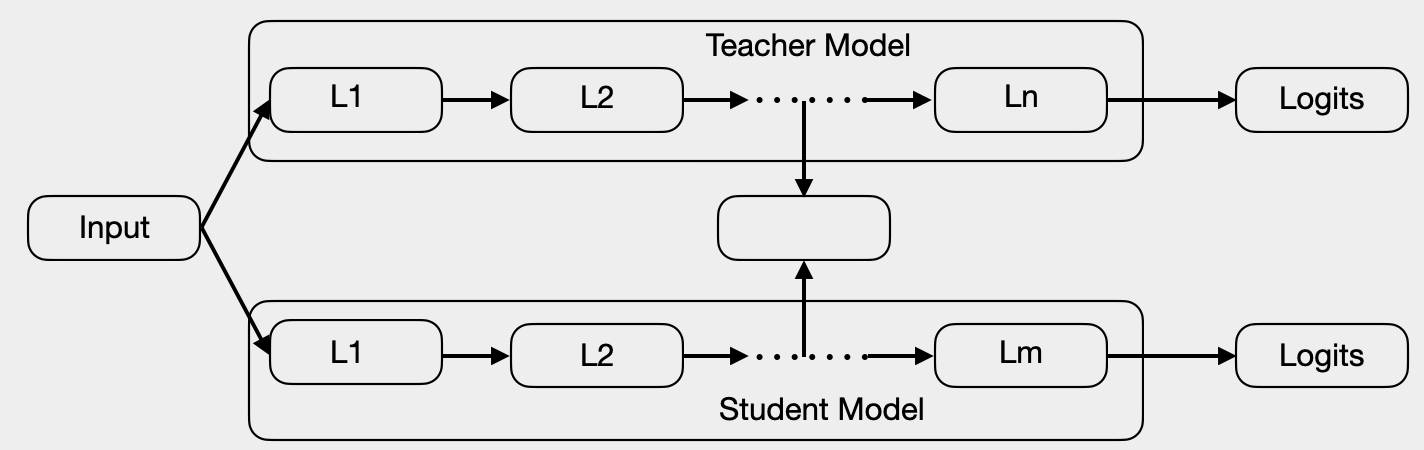}
\caption{Distillation based approaches}
\label{fig:distillation}
\end{figure}

Under the continued presupposition that the performance of the multimodal models is better in a high resource language. As a teacher model is learnt in this high resource language (usually English), the student model imitates this teacher model to learn task specific parameters to achieve a generalized performance. This is presented in Figure \ref{fig:distillation}.

Building on top of the monolingual multimodal models, \citet{raj-khan-etal-2021-towards-developing} use distillation methods to transfer the learning to multilingual and code-mixed scenarios for VQA. The teacher network is trained based on English LXMERT model, the parameters of which are used to train the student network to optimize for 2 mean squared errors and 2 binary cross entropy losses. The MSE are computed for the CLS token, object attention loss and the BCE losses are computed between the answer probability scores of teacher and student networks, and between the gold and the predicted answer from the student network.

\subsection{Pivoting}

\paragraph{Pivoting on the image: } \mbox{} \\
Multilingual multimodal datasets are often translated to different languages for the same visual context.
\citet{huang-etal-2021-multilingual} use a noise contrastive objective to the visually pivoted translation pairs between languages in inter-modal (i.e one language and visual), and intra-modal (i.e 2 languages and visual) ways. The goal of optimizing these objectives is to align the visual to the transcriptions in different languages. Prior to this, they perform pseudo visual pivoting \cite{huang-etal-2020-unsupervised-multimodal} motivated by  back translation to align multilingual spaces for the same image. Synthetic multilingual captions from the source image are used to reconstruct the synthetic captions from their corresponding translations and for translation of the paired captions. \citet{kadar-etal-2018-lessons} train a multilingual model to minimize the ranking loss updated for the prediction of image and caption of one language from caption in another language. A very similar approach with two pairwise ranking objectives scoring sentences and images and another scoring sentences in two different languages is also used by \citet{calixto-liu-2017-sentence}.
\citet{gella-etal-2017-image} optimize for a pivoted loss function on the image to bring gold description and image closer compared to other irrelevant captions using monolingual corpora from multiple languages. 

\paragraph{Alignment: } \mbox{} \\

\begin{figure}[tbh]
\centering
\includegraphics[trim=0cm 0cm 0cm 0cm,clip,width=0.4\linewidth]{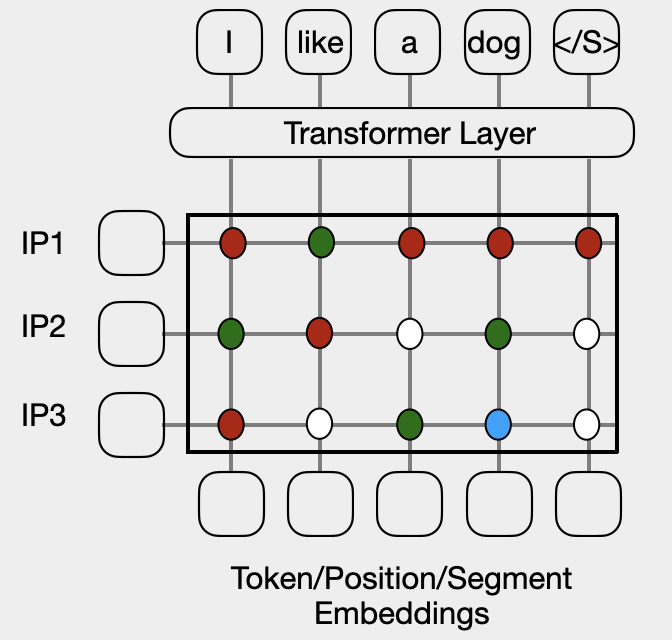}
\caption{Alignment based approaches}
\label{fig:alignment}
\end{figure}

Alignment is the method of forming a grouped plan to relate the languages and the modalities. 
Typically, this is achieved using an attention matrix. The attention based mehods are described in detailed after this method. Specifically, in Figure \ref{fig:alignment} the positioning matrix between the image (on the y-axis) and one of the languages (on the x-axis) to make predictions in the second language.

\citet{fei-etal-2021-cross} perform cross-lingual cross-modal pretraining with a unified framework using pretrained objectives adopted from prior studies including MLM (Masked Language Modeling) \cite{DBLP:conf/naacl/DevlinCLT19}, MRC (Masked region classification) \cite{DBLP:conf/aaai/LiDFGJ20, DBLP:conf/iclr/SuZCLLWD20}, TLM, CLTR (Cross-lingual text recovery), and CMTR (Cross-modal text recovery) \cite{DBLP:conf/emnlp/HuangLDGSJZ19}. Specifically, CLTR handles alignment of the parallel sentences in different languages by using bi-linear attention mechanism to compute an attended representation of the input sentence in the source language and its parallel sentence in a different language. Similarly, 
CMTR computes the alignment between word features and bounding box features by computing their bi-linear attention. 
Similar alignment techniques are also used to compute cross-lingual attention \cite{DBLP:conf/emnlp/GargPNP19}. \citet{nishihara-etal-2020-supervised} use a transformer based multimodal neural machine translation model with an image and source language sentence encoder and a target language decoder. The main addition to leverage cross-lingual information is derived by minimizing the cross entropy between one attention head of the multi-head cross-lingual attention  and the word alignments obtained using MGIZA \cite{DBLP:conf/emnlp/GargPNP19}. This cross-lingual alignment techniques such as XeroAlign \cite{DBLP:conf/acl/GrittaI21} and CrossAligner \cite{DBLP:conf/acl/GrittaHI22} is very prevalent in unimodal multilingual scenarios.
Introducing a diversity objective to explicitly capture different forms in the joint embedding space is an effective way to align terms in multiple langauges pivoted on an image \cite{huang-etal-2019-multi}. They perform multi-head attention to attend to different visual objects and the textual semantics in the caption with a margin-based diversity loss. Recently, \citet{DBLP:conf/cvpr/SurisEV22} present a modeling technique to learn aligned embedding space through a vision-based transitive relation across languages that learns an alignment model across languages if and only if the visuals associated with them are similar.

\paragraph{Attention: } \mbox{} \\

\citet{bagher-zadeh-etal-2020-cmu} use multimodal transformer \cite{DBLP:conf/acl/TsaiBLKMS19} composed of Conv1D and transformer to encode each modality to benchmark CMU-MOSEAS. Each langauge is independently modeled with the visual attributes without relying on cross-lingual information.
\citet{ramnath-etal-2021-worldly} use a co-attention mechanism to fuse selective information from the image and question for spoken visual question answering. Self-attention is performed on the speech signal and the question embeddings are used to query the attention on the image to answer a question from knowledge graphs represented by IaK \cite{DBLP:journals/corr/abs-2012-15484}. 
Instead of fusing the attended representation, \citet{shimizu-etal-2018-visual} replace the visual attention maps learnt from English to perform VQA in Japanese. The motivation for this idea is despite varying attention maps, their foci corresponding to the answer or the subject of the question have reasonable overlap across languages. This is followed by parallel coattention between visual and textual features. 
Similar cross attention on a single input language also shows improvements in multimodal machine translation in addition to multitasking with the auxiliary objective to construct a vision and language joint semantic embedding \cite{zhou-etal-2018-visual}.
Attention-based NMT frameworks  with attention on spatial features in the images are also used by combining multiple languages \cite{singh-etal-2021-multiple}. They define three types of mapping based on the source and target languages - many to one, one to many and many to many (where first and second indicate the number of source and target languages respectively). The many to one and many to many paradigms are cross-lingual where text from other languages is used during training. A similar approach of attending to the source sentence and the image is performed by \citet{imankulova-etal-2020-towards} for the task of simultaneous multimodal machine translation using wait-k approach. Instead of attention across languages, batching the data from different languages during training enables learn a multlingual representation \cite{gupta-etal-2020-unified}. Sharing this representation is also extended to attention based soft layer-sharing by attending over the encoder for each language and each layer fusing the modalities with bilinear attention. 

\paragraph{Other: } \mbox{} \\
\citet{karakanta-etal-2021-simultaneous} use speech translation based on an audio encoder and a text decoder using listen-and-translate \cite{DBLP:journals/corr/BerardPSB16} and direct foreign speech translation \cite{DBLP:conf/interspeech/WeissCJWC17}. These techniques are combined with the efficiency of the wait-k strategy \cite{DBLP:conf/acl/MaHXZLZZHLLWW19}. \citet{karakanta-etal-2020-must} present an NMT based model for subtitle generation of audio visual content using a transformer based seq2seq architecture for text only (visual features are not used in the model). 
\citet{liu-etal-2021-visually} set up baselines for MARVL on monolingual multimodal models including VL-BERT \cite{DBLP:conf/iclr/SuZCLLWD20}, VisualBERT \cite{DBLP:journals/corr/abs-1908-03557}, ViLBERT \cite{DBLP:conf/nips/LuBPL19}, LXMERT, and by extending English based multimodal models to multilingual scenarios. UNITER \cite{DBLP:conf/eccv/ChenLYK0G0020} is extended by initializing the text encoder with mBERT and XLM-R as mUNITER and xUNITER respectively. A similar approach is adopted by \citet{mitzalis-etal-2021-bertgen} to propose BERTGEN by fusing VL-BERT with M-BERT initialization. Specifically, it is demonstrated successfully for the task of MMT where unrolling is used as masking to create the next example and self attention is performed at every time step. 


\paragraph{Systems and Analysis} \mbox{} \\
\citet{akhlaghi-etal-2020-constructing} built LARA (Learning and Reading Assistant) which is an open source platform that converts plain texts into multimodal online versions. It involves semi-automatically tagging text, adding annotations, recording audio to highlight relevant information for suitable for language learners. Along similar lines, \citet{willemsen-etal-2018-context} also develop an L2 acquisition (L2TOR ITS) with a curriculum, state tracking and a template based NLG module for interaction. 
\citet{vilares-etal-2020-bringing} developed a description generator module for visually-impaired users to play 3 rogue-like games (The Inner Eye, The Accessible Dungeon, Hamsun’s Amulet). They follow a modular approach to content planning, micro-planning and surface realization of the NLG system proposed by \citet{DBLP:journals/nle/ReiterD97} using textual features at lexical, syntactic (POS tags), semantic (Multilingual Central Repository \cite{agirre2012multilingual}), discourse and pragmatic levels.  \citet{xu-etal-2020-xiaomingbot} built Xiaomingbot which is a software news reading robot with the capabilities of news generation, news translation into other languages, news reading along with avatar animation. The text for summarization is represented using BERT and a neural network with sliding window is used to generate smooth animations for the reader. \citet{poignant-etal-2016-camomile} built  a framework CAMOMILE client-server platform which is a collaborative annotation framework for collecting multimodal, multimedia, multilingual (3M) data. It is used to collect data for for MediaEval task with 20 annotators and upto 73k annotations. \citet{rinsche-2005-computer} presents LTC Communicator offering software web based multilingual support for vendors of international markets to support customers across countries interacting in multiple languages, which can be extended to information exchange via visuals of the products.

\noindent \textbf{Analysis}
\cite{moneglia-varvara-2020-annotation} perform analysis on IMAGACT \cite{DBLP:conf/lrec/MonegliaBFGKMP14} to understand the relation between thematic structure and semantic and lexical variation of action words. 

\section{Roadmap}
\label{sec:future}
Assimilating the takeaways from tasks and modeling for MultiX, we investigate the forefront of current challenges and promising directions.

\subsection{General Trends}

Generally speaking, the studies on MultiX has seen a paradigm shift with pretraining that demands a vast amount of data. Unlike monolingual processing that just needs raw data for self-supervised learning, the constraints of modalities and languages requires a degree of \textit{parallelness} in the data. Catering to this need, the field also obliged to using silver standard translated data to perform large scale modeling of MultiX compared to prior approaches of annotating gold standard multilingual data.

The base architectures or backbone models catering to the language aspect are mostly multilingual models such as mBART \cite{DBLP:journals/tacl/LiuGGLEGLZ20}, etc., Prior to the advent of pretrained multilingual models, most of the backbone architectures are CNN and LSTM based for visual and textual information respectively. Similarly, MFCC features are extracted to represent the speech modality. Some multilingual models also train on multiple languages together with an identifier token to uniquely identify to predict for that specific language \cite{mitzalis-etal-2021-bertgen}.

Overall, the field of multimodality seems to be extending a welcoming hand by  monolingual unimodal processing towards MultiX. However, the gap still remains owing to the inter-disciplinary topics of translation and multimodality. This creates an opportunity to progress in the field to build equitable technologies for all languages.

\subsection{Challenges and Directions}

\paragraph{Comparison to Unimodal approaches: }
From the findings of the shared task in Multimodal Machine Translation, \citet{barrault-etal-2018-findings} note that text only models are as competitive as multimodal models. While this is a common problem in monolingual multimodal cases, this observation is also an emerging trend in MultiX owing to the lack of strong underlying backbone architectures.
We encourage the community to scrupulously sub-select instances in the dataset containing concepts/words with ambiguity that enforces understanding of both modalities to predict the correct output. Maintaining the training data distributions with a wide coverage of high quality instances with this unimodal or monolingual ambiguity is promising to better compare unimodal and multimodal models. 

\begin{quote}
\textit{Direction: Designing tasks with ambiguous relations where a single modality curtails quality predictions are critical.
}
\end{quote}

\begin{figure}[t!]
\centering
\includegraphics[trim=7.6cm 12cm 7.6cm 12cm,clip,width=0.99\linewidth]{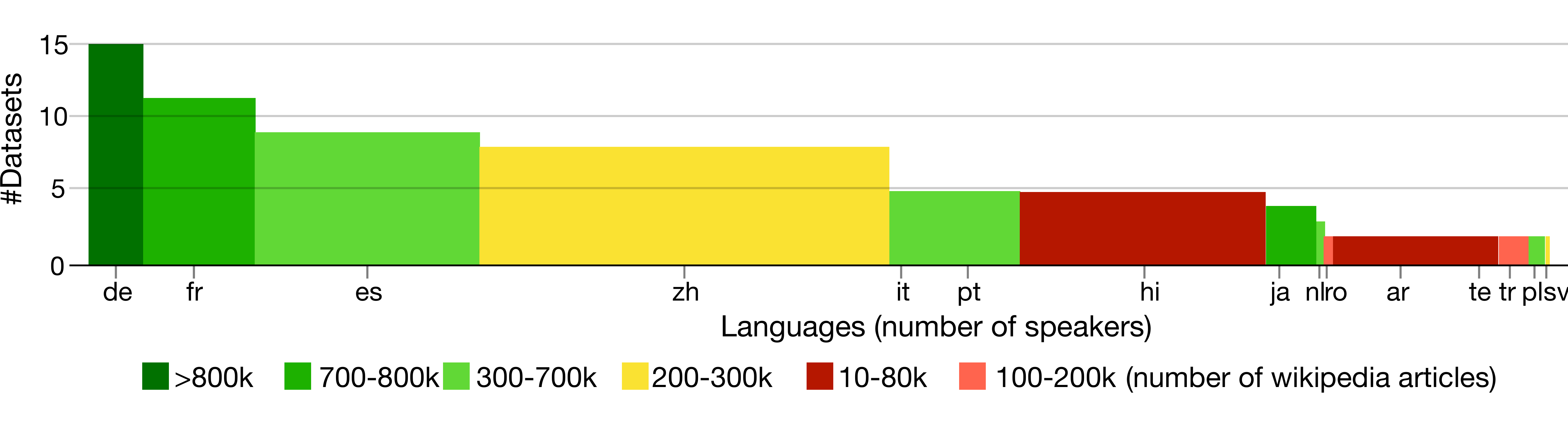}
\caption{Distribution of MultiX datasets for languages with  the number of speakers and Wikipedia articles}
\label{fig:distribution_langs}
\end{figure}

\paragraph{Few shot Cross-lingual Transfer: }
\citet{DBLP:conf/acl/BlasiAN22} presented disparities in NLP technologies due to the societal and academic factors. In a similar spirit, we present the demographic utility of MultiX datasets in Figure  \ref{fig:distribution_langs}, which proliferate to modeling resources. We can observe that the digital presence in terms of the number of Wikipedia articles is correlated with datasets built in these languages irrespective of the number of speakers. In this sub-optimal reality, incentivizing research in underrepresented and endangered languages \cite{DBLP:conf/coling/Bird20} is practical with zero or few-shot transfer. \citet{DBLP:conf/emnlp/LauscherRVG20} study the relationship between the success of transfer with varying levels of tasks, language proximity, amount of target language data. concluding that few shot finetuning has significant boost over zero-shot transfer. 
A common approach for a multilingual multimodal model, finetune the model for the task on a high-resourced language, perform 0-shot or few-shot finetuning on the low-resourced languages.

\begin{quote}
\textit{Direction: Leveraging shared linguistic units grounded visually at syntactic \& lexical levels to maximize supervision, assists transfer from high resourced languages.
}
\end{quote}

\paragraph{Multi vs Bilingual: } 
Multilingually aware end-to-end systems are better for error propagation \cite{DBLP:conf/emnlp/ZhuWWZZWZ19, DBLP:conf/ijcnlp/XuZSZH20} in unimodal scenarios.
Various studies demonstrate that multilingual training improves performance over bilingual training, or by extrapolation, monolingual training \cite{kadar-etal-2018-lessons}. 
Similarly, the asymmetric loss \cite{DBLP:journals/corr/VendrovKFU15} in pivoted models by \citet{gella-etal-2017-image} suggest that multilingual information sharing is useful. 
This observation motivates that when collecting multimodal data for a new language, it is beneficial to collect for the same images with existing data in another language \cite{chandu2021grounding} to exploit caption-to-caption prediction objectives along with image prediction objectives. 
However, we need to be cognizant of the abberations where \citet{ku-etal-2020-room} were not able to directly improve performance by multilingual models, so they take the task diversity as an additional dimension for multitasking to improve multilingual navigation. 

\begin{quote}
\textit{Direction: Multilingual learning has the potential to enhance MultiX, which benefits from parallel/pseudo-parallel data across multiple languages.
}
\end{quote}


\paragraph{Curse of Multilinguality: }
Monolingual performance with fixed model capacity trained on multiple languages starts weakening with the addition of new languages, known as the curse of multilinguality \cite{DBLP:conf/acl/ConneauKGCWGGOZ20}. This problem of locked capacity can be fixed with augmentation of model parameters and data.
First, augmenting data can be noisy supervision with silver  annotations or unsupervised data.
Second, augmenting models involves adding language specific adapters \cite{DBLP:conf/emnlp/PfeifferVGR20} or multilingual tokenizers \cite{DBLP:conf/acl/RustPVRG20}. 
The language and task specific adapters are tuned separately and the adapter of the source language is replaced with that of the target language in inference for the same multimodal model. However, a very recent work by \citet{DBLP:journals/corr/abs-2205-03983} showed that the curse of multilinguality does not hold when there are several more languages of the order of thousands based on experiments conducted where the authors do not observe this interference effect on a single multilingual model.


\begin{quote}
\textit{Direction: Adapter based tuning is a parameter-efficient way to ensure continual cross-lingual transfer for pretrained  multimodal models.
}
\end{quote}

\paragraph{Translationese artifacts: } While translating existing monolingual multimodal resources offers developing parallel data effective for cross-lingual learning, they often fall prey to translationese without explicit caution.
\textit{Translationese} \cite{DBLP:conf/acl/KoppelO11} is characterized as the language style resulting from a translator attempting to closely replicate the properties of source text thereby heavily biasing it.  
Translations are often error-proned with low-resourced languages and colloquial usage of terms adversely impacting evaluation \cite{DBLP:conf/emnlp/GrahamHK20}. Partial translations can help mitigate this.  
Studies show that this language style also adversely impacts the evaluation \cite{DBLP:conf/emnlp/GrahamHK20}. 
First, translating continuous monolingual chunks instead of full sentences inherently models cross-lingual contextual information. A multimodal code-switched stream is explored in  pretraining M3P \cite{DBLP:conf/cvpr/NiHSCBW0D21} and in unimodal tasks \cite{DBLP:conf/ijcai/QinN0C20, DBLP:journals/corr/abs-2103-07792} with random replacements.
Second, a hybrid approach is translating approximated or delexicalized templates and then filling in with regionally relevant tokens in the translated template \cite{DBLP:conf/acl/DingHBMJSM22}, which can also help reduce human effort and cost for annotations. 
Therefore, translating partial utterances grounded in entities,  context, etc., mitigates translating entire source annotation, thereby reducing bias.

\begin{quote}
\textit{Direction: Minimizing source language artifacts to ground cross-lingual annotations curbs translationese.
}
\end{quote}

\paragraph{Language Pivoting: } As we studied in \S  \ref{sec:methods}, pivoting is primarily done on the image. Realistically, availability of data for the axes, (i) paired images and text, compared to other languages, 
and (ii) paired translations for English is relatively more common.
This imminently rises the opportunity to pivot on a resource rich languages common between both axes. \citet{DBLP:conf/eccv/GuJCW18} studied this way of unpaired captioning with language pivoting performing machine translation and autoencoding with paired data to in turn perform captioning without paired data. However, selecting English as the pivot language is not the best choice \cite{DBLP:conf/acl/AnastasopoulosN20} and this selection based on typological distance is still unsolved.

\begin{quote}
\textit{Direction: Pivoting on a resource rich common language is promising for unpaired learning with a preference of common language proximal to target language.
}
\end{quote}

\paragraph{Expansive X: }
Societal grounding influences regional language usage, making the dialects, vernaculars, accents, idiolects and other colloquial variants different from the language itself. In addition, catering to domain shifts \cite{DBLP:conf/coling/RamponiP20} in languages encourages out-of-distribution generalization. Expanding to these ever-evolving variants of Xs demands reliable evaluations in an ever-evolving area that we need to be cognizant of.

\begin{quote}
\textit{Direction: Sustainable research demands developing comparable evaluation framework leaning into the cultural differences and morphological richness.
}
\end{quote}

With the rapid development of multimodal and multilingual axes independently, it is now more critical than ever to build multimodal models accommodative of several languages. With this broad goal, this paper presents categories of tasks, datasets, and methods along with sketching out existing challenges and a roadmap ahead.

\section*{Limitations}
This paper studies unifying two of the important dimensions in contexts. However, the dimensions are not limited to the scope of multiple languages or modalities. 
First, on top of the variants discussed in \S \ref{sec:future} in \textit{``Expansive X"}, domains are an important category of language usage that warrant specialized care to ensure competitive performance. This survey does not broadly include the domains within \textit{X}, however, it is imperative to understand the effects of domains and distributional shifts as we progress on the dimensions of languages and modalities. Note that with the existing two dimensions, the distributional shift can occur within a language or the visual sources, thereby coupling the complexity of the space. Second, the majority of the work covered in this paper does not address underrepresented languages due to lack of existing literature. However, we need to take inspiration from the growing research on low resourced languages \cite{DBLP:conf/naacl/HedderichLASK21} to ensure we learn from what works best in such unimodal cases to coalesce with the multimodal techniques. To deliver an equitable societal impact of language technologies, it is important to address this long-tail of low-resourced languages as well. 

\section*{Broader Impact}
The fields of multilingual NLP and multimodality are racing with the  vibrant and fierce contributions from our community. However, these resources and methods are not directly trivially extensible to languages other than the one they are developed in, which is English in most cases. Hence, it is important essential to retrospect the resources and techniques that researchers proposed in MultiX to assimilate the takeaways to consequently build informed and equitable models. 
Extending existing vision-and-language resources to multiple languages ensures not only the re-usability of the work done so far but also builds an ecosystem to compare the scalability and robustness of the models for continued research.
We hope this work can be used as an index into the wealth of scattered resources unifying multimodality and multilinguality  for researchers in both these fields 
to be cognizant of the challenges and promising directions to invest upon while making modeling choices.

\section*{Ethical Considerations}
We do not presage any immediate ethical concerns arising directly from our work on surveying the landscape of MultiX.  
The taxonomy of datasets and techniques focuses on multilingual multimodality i.e., MultiX, which does not address undesired and harmful biases inherently present in the datasets. Moreover, the cultural norms dictate the linguistic acceptability of translated resources. Therefore the proposed translation based weak supervision methods need to carefully take into account cultural qualifications specific to each language and region.
\newpage

\appendix

\section{Acronyms}

Here are some of the acronyms discussed in the paper along with their expansions.

\begin{itemize}
\item ASR: Automatic Speech Recognition
\item BCE: Binary Cross Entropy
\item COCO: Common Objects in Context
\item CCA: Canonical Correlation Analysis
\item CLIP: Contrastive Language–Image Pre-training
\item CMU-MOSEAS: CMU Multimodal Opinion Sentiment, Emotions and Attributes
\item FM-IQA: Freestyle Multilingual Image Question Answering
\item IaK: Image as Knowledge
\item IPA: International Phonetic Alphabet
\item MARVL: Multicultural Reasoning over Vision and Language
\item MFCC: Mel Frequency Cepstral Coefficient
\item MMT: Multimodal Machine Translation
\item MSE: Mean squared error
\item MuST-C: Multilingual Speech Translation Corpus
\item MuST-Cinema: a Multilingual Speech Translation Cinema
\item NMT: Neural Machine Translation
\item OpenCLIR: Open Cross Language Information Retrieval
\item OpenNMT: Open Neural Machine Traslation
\item QA: Question Answering
\item RNN: Recurrent Neural Network
\item seq2seq: sequence to sequence
\item TED: Technology, Entertainment, and Design
\item TTS: Text-to-Speech
\item VQA: Visual Question Answering
\end{itemize}

\vskip 0.2in
\bibliography{sample}

\begin{thebibliography}{137}
\providecommand{\natexlab}[1]{#1}
\providecommand{\url}[1]{\texttt{#1}}
\expandafter\ifx\csname urlstyle\endcsname\relax
  \providecommand{\doi}[1]{doi: #1}\else
  \providecommand{\doi}{doi: \begingroup \urlstyle{rm}\Url}\fi

\bibitem[Afli et~al.(2017)Afli, Lohar, and Way]{afli-etal-2017-multinews}
Haithem Afli, Pintu Lohar, and Andy Way.
\newblock {M}ulti{N}ews: A web collection of an aligned multimodal and
  multilingual corpus.
\newblock In \emph{Proceedings of the First Workshop on Curation and
  Applications of Parallel and Comparable Corpora}, pages 11--15, Taipei,
  Taiwan, November 2017. Asian Federation of Natural Language Processing.
\newblock URL \url{https://aclanthology.org/W17-5602}.

\bibitem[Agirre et~al.(2012)Agirre, Laparra, Rigau, and
  Donostia]{agirre2012multilingual}
Aitor~Gonz{\'a}lez Agirre, Egoitz Laparra, German Rigau, and Basque~Country
  Donostia.
\newblock Multilingual central repository version 3.0: upgrading a very large
  lexical knowledge base.
\newblock In \emph{GWC 2012 6th International Global Wordnet Conference}, page
  118, 2012.

\bibitem[Akhlaghi et~al.(2020)Akhlaghi, Bedi, Bekta{\c{s}}, Berthelsen,
  Butterweck, Chua, Cucchiarin, Eryigit, Gerlach, Habibi, Ni~Chiarain, Rayner,
  Steingrimsson, and Strik]{akhlaghi-etal-2020-constructing}
Elham Akhlaghi, Branislav Bedi, Fatih Bekta{\c{s}}, Harald Berthelsen, Matthias
  Butterweck, Cathy Chua, Catia Cucchiarin, Gulsen Eryigit, Johanna Gerlach,
  Hanieh Habibi, Neasa Ni~Chiarain, Manny Rayner, Steinor Steingrimsson, and
  Helmer Strik.
\newblock Constructing multimodal language learner texts using {LARA}:
  Experiences with nine languages.
\newblock In \emph{Proceedings of the 12th Language Resources and Evaluation
  Conference}, pages 323--331, Marseille, France, May 2020. European Language
  Resources Association.
\newblock ISBN 979-10-95546-34-4.
\newblock URL \url{https://aclanthology.org/2020.lrec-1.40}.

\bibitem[Alberts et~al.(2021)Alberts, Huang, Deshpande, Liu, Cho, Vania, and
  Calixto]{alberts-etal-2021-visualsem}
Houda Alberts, Ningyuan Huang, Yash Deshpande, Yibo Liu, Kyunghyun Cho, Clara
  Vania, and Iacer Calixto.
\newblock {V}isual{S}em: a high-quality knowledge graph for vision and
  language.
\newblock In \emph{Proceedings of the 1st Workshop on Multilingual
  Representation Learning}, pages 138--152, Punta Cana, Dominican Republic,
  November 2021. Association for Computational Linguistics.
\newblock \doi{10.18653/v1/2021.mrl-1.13}.
\newblock URL \url{https://aclanthology.org/2021.mrl-1.13}.

\bibitem[Anastasopoulos and Neubig(2020)]{DBLP:conf/acl/AnastasopoulosN20}
Antonios Anastasopoulos and Graham Neubig.
\newblock Should all cross-lingual embeddings speak english?
\newblock In Dan Jurafsky, Joyce Chai, Natalie Schluter, and Joel~R. Tetreault,
  editors, \emph{Proceedings of the 58th Annual Meeting of the Association for
  Computational Linguistics, {ACL} 2020, Online, July 5-10, 2020}, pages
  8658--8679. Association for Computational Linguistics, 2020.
\newblock \doi{10.18653/v1/2020.acl-main.766}.
\newblock URL \url{https://doi.org/10.18653/v1/2020.acl-main.766}.

\bibitem[Anderson et~al.(2018)Anderson, Wu, Teney, Bruce, Johnson,
  S{\"{u}}nderhauf, Reid, Gould, and van~den
  Hengel]{DBLP:conf/cvpr/AndersonWTB0S0G18}
Peter Anderson, Qi~Wu, Damien Teney, Jake Bruce, Mark Johnson, Niko
  S{\"{u}}nderhauf, Ian~D. Reid, Stephen Gould, and Anton van~den Hengel.
\newblock Vision-and-language navigation: Interpreting visually-grounded
  navigation instructions in real environments.
\newblock In \emph{2018 {IEEE} Conference on Computer Vision and Pattern
  Recognition, {CVPR} 2018, Salt Lake City, UT, USA, June 18-22, 2018}, pages
  3674--3683. Computer Vision Foundation / {IEEE} Computer Society, 2018.
\newblock \doi{10.1109/CVPR.2018.00387}.
\newblock URL
  \url{http://openaccess.thecvf.com/content\_cvpr\_2018/html/Anderson\_Vision-and-Language\_Navigation\_Interpreting\_CVPR\_2018\_paper.html}.

\bibitem[Arora et~al.(2020)Arora, Shterionov, Moriya, Kaushik, Dzendzik, and
  Jones]{arora-etal-2020-investigative}
Piyush Arora, Dimitar Shterionov, Yasufumi Moriya, Abhishek Kaushik, Daria
  Dzendzik, and Gareth Jones.
\newblock An investigative study of multi-modal cross-lingual retrieval.
\newblock In \emph{Proceedings of the workshop on Cross-Language Search and
  Summarization of Text and Speech (CLSSTS2020)}, pages 58--67, Marseille,
  France, May 2020. European Language Resources Association.
\newblock ISBN 979-10-95546-55-9.
\newblock URL \url{https://aclanthology.org/2020.clssts-1.10}.

\bibitem[Bagher~Zadeh et~al.(2020)Bagher~Zadeh, Cao, Hessner, Liang, Poria, and
  Morency]{bagher-zadeh-etal-2020-cmu}
AmirAli Bagher~Zadeh, Yansheng Cao, Simon Hessner, Paul~Pu Liang, Soujanya
  Poria, and Louis-Philippe Morency.
\newblock {CMU}-{MOSEAS}: A multimodal language dataset for {S}panish,
  {P}ortuguese, {G}erman and {F}rench.
\newblock In \emph{Proceedings of the 2020 Conference on Empirical Methods in
  Natural Language Processing (EMNLP)}, pages 1801--1812, Online, November
  2020. Association for Computational Linguistics.
\newblock \doi{10.18653/v1/2020.emnlp-main.141}.
\newblock URL \url{https://aclanthology.org/2020.emnlp-main.141}.

\bibitem[Baltrusaitis et~al.(2019)Baltrusaitis, Ahuja, and
  Morency]{DBLP:journals/pami/BaltrusaitisAM19}
Tadas Baltrusaitis, Chaitanya Ahuja, and Louis{-}Philippe Morency.
\newblock Multimodal machine learning: {A} survey and taxonomy.
\newblock \emph{{IEEE} Trans. Pattern Anal. Mach. Intell.}, 41\penalty0
  (2):\penalty0 423--443, 2019.
\newblock \doi{10.1109/TPAMI.2018.2798607}.
\newblock URL \url{https://doi.org/10.1109/TPAMI.2018.2798607}.

\bibitem[Bapna et~al.(2022)Bapna, Caswell, Kreutzer, Firat, van Esch, Siddhant,
  Niu, Baljekar, Garcia, Macherey, Breiner, Axelrod, Riesa, Cao, Chen,
  Macherey, Krikun, Wang, Gutkin, Shah, Huang, Chen, Wu, and
  Hughes]{DBLP:journals/corr/abs-2205-03983}
Ankur Bapna, Isaac Caswell, Julia Kreutzer, Orhan Firat, Daan van Esch, Aditya
  Siddhant, Mengmeng Niu, Pallavi Baljekar, Xavier Garcia, Wolfgang Macherey,
  Theresa Breiner, Vera Axelrod, Jason Riesa, Yuan Cao, Mia~Xu Chen, Klaus
  Macherey, Maxim Krikun, Pidong Wang, Alexander Gutkin, Apurva Shah, Yanping
  Huang, Zhifeng Chen, Yonghui Wu, and Macduff Hughes.
\newblock Building machine translation systems for the next thousand languages.
\newblock \emph{CoRR}, abs/2205.03983, 2022.
\newblock \doi{10.48550/arXiv.2205.03983}.
\newblock URL \url{https://doi.org/10.48550/arXiv.2205.03983}.

\bibitem[Barrault et~al.(2018)Barrault, Bougares, Specia, Lala, Elliott, and
  Frank]{barrault-etal-2018-findings}
Lo{\"\i}c Barrault, Fethi Bougares, Lucia Specia, Chiraag Lala, Desmond
  Elliott, and Stella Frank.
\newblock Findings of the third shared task on multimodal machine translation.
\newblock In \emph{Proceedings of the Third Conference on Machine Translation:
  Shared Task Papers}, pages 304--323, Belgium, Brussels, October 2018.
  Association for Computational Linguistics.
\newblock \doi{10.18653/v1/W18-6402}.
\newblock URL \url{https://aclanthology.org/W18-6402}.

\bibitem[Berard et~al.(2016)Berard, Pietquin, Servan, and
  Besacier]{DBLP:journals/corr/BerardPSB16}
Alexandre Berard, Olivier Pietquin, Christophe Servan, and Laurent Besacier.
\newblock Listen and translate: {A} proof of concept for end-to-end
  speech-to-text translation.
\newblock \emph{CoRR}, abs/1612.01744, 2016.
\newblock URL \url{http://arxiv.org/abs/1612.01744}.

\bibitem[Bird(2020)]{DBLP:conf/coling/Bird20}
Steven Bird.
\newblock Decolonising speech and language technology.
\newblock In Donia Scott, N{\'{u}}ria Bel, and Chengqing Zong, editors,
  \emph{Proceedings of the 28th International Conference on Computational
  Linguistics, {COLING} 2020, Barcelona, Spain (Online), December 8-13, 2020},
  pages 3504--3519. International Committee on Computational Linguistics, 2020.
\newblock \doi{10.18653/v1/2020.coling-main.313}.
\newblock URL \url{https://doi.org/10.18653/v1/2020.coling-main.313}.

\bibitem[Black(2019)]{DBLP:conf/icassp/Black19}
Alan~W. Black.
\newblock {CMU} wilderness multilingual speech dataset.
\newblock In \emph{{IEEE} International Conference on Acoustics, Speech and
  Signal Processing, {ICASSP} 2019, Brighton, United Kingdom, May 12-17, 2019},
  pages 5971--5975. {IEEE}, 2019.
\newblock \doi{10.1109/ICASSP.2019.8683536}.
\newblock URL \url{https://doi.org/10.1109/ICASSP.2019.8683536}.

\bibitem[Blasi et~al.(2022)Blasi, Anastasopoulos, and
  Neubig]{DBLP:conf/acl/BlasiAN22}
Dami{\'{a}}n~E. Blasi, Antonios Anastasopoulos, and Graham Neubig.
\newblock Systematic inequalities in language technology performance across the
  world's languages.
\newblock In Smaranda Muresan, Preslav Nakov, and Aline Villavicencio, editors,
  \emph{Proceedings of the 60th Annual Meeting of the Association for
  Computational Linguistics (Volume 1: Long Papers), {ACL} 2022, Dublin,
  Ireland, May 22-27, 2022}, pages 5486--5505. Association for Computational
  Linguistics, 2022.
\newblock URL \url{https://aclanthology.org/2022.acl-long.376}.

\bibitem[Bugliarello et~al.(2022)Bugliarello, Liu, Pfeiffer, Reddy, Elliott,
  Ponti, and Vulic]{DBLP:conf/icml/Bugliarello0PRE22}
Emanuele Bugliarello, Fangyu Liu, Jonas Pfeiffer, Siva Reddy, Desmond Elliott,
  Edoardo~Maria Ponti, and Ivan Vulic.
\newblock {IGLUE:} {A} benchmark for transfer learning across modalities,
  tasks, and languages.
\newblock In Kamalika Chaudhuri, Stefanie Jegelka, Le~Song, Csaba
  Szepesv{\'{a}}ri, Gang Niu, and Sivan Sabato, editors, \emph{International
  Conference on Machine Learning, {ICML} 2022, 17-23 July 2022, Baltimore,
  Maryland, {USA}}, volume 162 of \emph{Proceedings of Machine Learning
  Research}, pages 2370--2392. {PMLR}, 2022.
\newblock URL \url{https://proceedings.mlr.press/v162/bugliarello22a.html}.

\bibitem[Burger et~al.(2003)Burger, Costantini, and
  Pianesi]{burger-etal-2003-communicative}
Susanne Burger, Erica Costantini, and Fabio Pianesi.
\newblock Communicative strategies and patterns of multimodal integration in a
  speech-to-speech translation system.
\newblock In \emph{Proceedings of Machine Translation Summit IX: Papers}, New
  Orleans, USA, September 23-27 2003.
\newblock URL \url{https://aclanthology.org/2003.mtsummit-papers.5}.

\bibitem[Caglayan et~al.(2021)Caglayan, Kuyu, Amac, Madhyastha, Erdem, Erdem,
  and Specia]{caglayan-etal-2021-cross}
Ozan Caglayan, Menekse Kuyu, Mustafa~Sercan Amac, Pranava Madhyastha, Erkut
  Erdem, Aykut Erdem, and Lucia Specia.
\newblock Cross-lingual visual pre-training for multimodal machine translation.
\newblock In \emph{Proceedings of the 16th Conference of the European Chapter
  of the Association for Computational Linguistics: Main Volume}, pages
  1317--1324, Online, April 2021. Association for Computational Linguistics.
\newblock \doi{10.18653/v1/2021.eacl-main.112}.
\newblock URL \url{https://aclanthology.org/2021.eacl-main.112}.

\bibitem[Calixto and Liu(2017{\natexlab{a}})]{calixto-liu-2017-incorporating}
Iacer Calixto and Qun Liu.
\newblock Incorporating global visual features into attention-based neural
  machine translation.
\newblock In \emph{Proceedings of the 2017 Conference on Empirical Methods in
  Natural Language Processing}, pages 992--1003, Copenhagen, Denmark, September
  2017{\natexlab{a}}. Association for Computational Linguistics.
\newblock \doi{10.18653/v1/D17-1105}.
\newblock URL \url{https://aclanthology.org/D17-1105}.

\bibitem[Calixto and Liu(2017{\natexlab{b}})]{calixto-liu-2017-sentence}
Iacer Calixto and Qun Liu.
\newblock Sentence-level multilingual multi-modal embedding for natural
  language processing.
\newblock In \emph{Proceedings of the International Conference Recent Advances
  in Natural Language Processing, {RANLP} 2017}, pages 139--148, Varna,
  Bulgaria, September 2017{\natexlab{b}}. INCOMA Ltd.
\newblock \doi{10.26615/978-954-452-049-6_020}.
\newblock URL \url{https://doi.org/10.26615/978-954-452-049-6_020}.

\bibitem[Chandu et~al.(2021)Chandu, Bisk, and Black]{chandu2021grounding}
Khyathi~Raghavi Chandu, Yonatan Bisk, and Alan~W Black.
\newblock Grounding ‘grounding’in nlp.
\newblock In \emph{Findings of the Association for Computational Linguistics:
  ACL-IJCNLP 2021}, pages 4283--4305, 2021.

\bibitem[Chen et~al.(2020)Chen, Li, Yu, Kholy, Ahmed, Gan, Cheng, and
  Liu]{DBLP:conf/eccv/ChenLYK0G0020}
Yen{-}Chun Chen, Linjie Li, Licheng Yu, Ahmed~El Kholy, Faisal Ahmed, Zhe Gan,
  Yu~Cheng, and Jingjing Liu.
\newblock {UNITER:} universal image-text representation learning.
\newblock In Andrea Vedaldi, Horst Bischof, Thomas Brox, and Jan{-}Michael
  Frahm, editors, \emph{Computer Vision - {ECCV} 2020 - 16th European
  Conference, Glasgow, UK, August 23-28, 2020, Proceedings, Part {XXX}}, volume
  12375 of \emph{Lecture Notes in Computer Science}, pages 104--120. Springer,
  2020.
\newblock \doi{10.1007/978-3-030-58577-8\_7}.
\newblock URL \url{https://doi.org/10.1007/978-3-030-58577-8\_7}.

\bibitem[Conneau and Lample(2019)]{DBLP:conf/nips/ConneauL19}
Alexis Conneau and Guillaume Lample.
\newblock Cross-lingual language model pretraining.
\newblock In Hanna~M. Wallach, Hugo Larochelle, Alina Beygelzimer, Florence
  d'Alch{\'{e}}{-}Buc, Emily~B. Fox, and Roman Garnett, editors, \emph{Advances
  in Neural Information Processing Systems 32: Annual Conference on Neural
  Information Processing Systems 2019, NeurIPS 2019, December 8-14, 2019,
  Vancouver, BC, Canada}, pages 7057--7067, 2019.
\newblock URL
  \url{https://proceedings.neurips.cc/paper/2019/hash/c04c19c2c2474dbf5f7ac4372c5b9af1-Abstract.html}.

\bibitem[Conneau et~al.(2020)Conneau, Khandelwal, Goyal, Chaudhary, Wenzek,
  Guzm{\'{a}}n, Grave, Ott, Zettlemoyer, and
  Stoyanov]{DBLP:conf/acl/ConneauKGCWGGOZ20}
Alexis Conneau, Kartikay Khandelwal, Naman Goyal, Vishrav Chaudhary, Guillaume
  Wenzek, Francisco Guzm{\'{a}}n, Edouard Grave, Myle Ott, Luke Zettlemoyer,
  and Veselin Stoyanov.
\newblock Unsupervised cross-lingual representation learning at scale.
\newblock In Dan Jurafsky, Joyce Chai, Natalie Schluter, and Joel~R. Tetreault,
  editors, \emph{Proceedings of the 58th Annual Meeting of the Association for
  Computational Linguistics, {ACL} 2020, Online, July 5-10, 2020}, pages
  8440--8451. Association for Computational Linguistics, 2020.
\newblock \doi{10.18653/v1/2020.acl-main.747}.
\newblock URL \url{https://doi.org/10.18653/v1/2020.acl-main.747}.

\bibitem[Dabre et~al.(2020)Dabre, Chu, and
  Kunchukuttan]{DBLP:journals/csur/DabreCK20}
Raj Dabre, Chenhui Chu, and Anoop Kunchukuttan.
\newblock A survey of multilingual neural machine translation.
\newblock \emph{{ACM} Comput. Surv.}, 53\penalty0 (5):\penalty0 99:1--99:38,
  2020.
\newblock \doi{10.1145/3406095}.
\newblock URL \url{https://doi.org/10.1145/3406095}.

\bibitem[de~Melo and Weikum(2010)]{de-melo-weikum-2010-providing}
Gerard de~Melo and Gerhard Weikum.
\newblock Providing multilingual, multimodal answers to lexical database
  queries.
\newblock In \emph{Proceedings of the Seventh International Conference on
  Language Resources and Evaluation ({LREC}'10)}, Valletta, Malta, May 2010.
  European Language Resources Association (ELRA).
\newblock URL
  \url{http://www.lrec-conf.org/proceedings/lrec2010/pdf/312_Paper.pdf}.

\bibitem[Devlin et~al.(2019)Devlin, Chang, Lee, and
  Toutanova]{DBLP:conf/naacl/DevlinCLT19}
Jacob Devlin, Ming{-}Wei Chang, Kenton Lee, and Kristina Toutanova.
\newblock {BERT:} pre-training of deep bidirectional transformers for language
  understanding.
\newblock In Jill Burstein, Christy Doran, and Thamar Solorio, editors,
  \emph{Proceedings of the 2019 Conference of the North American Chapter of the
  Association for Computational Linguistics: Human Language Technologies,
  {NAACL-HLT} 2019, Minneapolis, MN, USA, June 2-7, 2019, Volume 1 (Long and
  Short Papers)}, pages 4171--4186. Association for Computational Linguistics,
  2019.
\newblock \doi{10.18653/v1/n19-1423}.
\newblock URL \url{https://doi.org/10.18653/v1/n19-1423}.

\bibitem[Ding et~al.(2022)Ding, Hu, Bing, Mahani, Joty, Si, and
  Miao]{DBLP:conf/acl/DingHBMJSM22}
Bosheng Ding, Junjie Hu, Lidong Bing, Mahani~Aljunied Mahani, Shafiq~R. Joty,
  Luo Si, and Chunyan Miao.
\newblock Globalwoz: Globalizing multiwoz to develop multilingual task-oriented
  dialogue systems.
\newblock In Smaranda Muresan, Preslav Nakov, and Aline Villavicencio, editors,
  \emph{Proceedings of the 60th Annual Meeting of the Association for
  Computational Linguistics (Volume 1: Long Papers), {ACL} 2022, Dublin,
  Ireland, May 22-27, 2022}, pages 1639--1657. Association for Computational
  Linguistics, 2022.
\newblock URL \url{https://aclanthology.org/2022.acl-long.115}.

\bibitem[Elliott et~al.(2016)Elliott, Frank, Sima'an, and
  Specia]{DBLP:conf/acl/ElliottFSS16}
Desmond Elliott, Stella Frank, Khalil Sima'an, and Lucia Specia.
\newblock Multi30k: Multilingual english-german image descriptions.
\newblock In \emph{Proceedings of the 5th Workshop on Vision and Language,
  hosted by the 54th Annual Meeting of the Association for Computational
  Linguistics, VL@ACL 2016, August 12, Berlin, Germany}. The Association for
  Computer Linguistics, 2016.
\newblock \doi{10.18653/v1/w16-3210}.
\newblock URL \url{https://doi.org/10.18653/v1/w16-3210}.

\bibitem[Elliott et~al.(2017)Elliott, Frank, Barrault, Bougares, and
  Specia]{DBLP:conf/wmt/ElliottFBBS17}
Desmond Elliott, Stella Frank, Lo{\"{\i}}c Barrault, Fethi Bougares, and Lucia
  Specia.
\newblock Findings of the second shared task on multimodal machine translation
  and multilingual image description.
\newblock In Ondrej Bojar, Christian Buck, Rajen Chatterjee, Christian
  Federmann, Yvette Graham, Barry Haddow, Matthias Huck, Antonio
  Jimeno{-}Yepes, Philipp Koehn, and Julia Kreutzer, editors, \emph{Proceedings
  of the Second Conference on Machine Translation, {WMT} 2017, Copenhagen,
  Denmark, September 7-8, 2017}, pages 215--233. Association for Computational
  Linguistics, 2017.
\newblock \doi{10.18653/v1/w17-4718}.
\newblock URL \url{https://doi.org/10.18653/v1/w17-4718}.

\bibitem[Fei et~al.(2021)Fei, Yu, and Li]{fei-etal-2021-cross}
Hongliang Fei, Tan Yu, and Ping Li.
\newblock Cross-lingual cross-modal pretraining for multimodal retrieval.
\newblock In \emph{Proceedings of the 2021 Conference of the North American
  Chapter of the Association for Computational Linguistics: Human Language
  Technologies}, pages 3644--3650, Online, June 2021. Association for
  Computational Linguistics.
\newblock \doi{10.18653/v1/2021.naacl-main.285}.
\newblock URL \url{https://aclanthology.org/2021.naacl-main.285}.

\bibitem[Fu et~al.(2017)Fu, Lee, Bansal, and Berg]{fu-etal-2017-video}
Cheng-Yang Fu, Joon Lee, Mohit Bansal, and Alexander Berg.
\newblock Video highlight prediction using audience chat reactions.
\newblock In \emph{Proceedings of the 2017 Conference on Empirical Methods in
  Natural Language Processing}, pages 972--978, Copenhagen, Denmark, September
  2017. Association for Computational Linguistics.
\newblock \doi{10.18653/v1/D17-1102}.
\newblock URL \url{https://aclanthology.org/D17-1102}.

\bibitem[Funaki and Nakayama(2015)]{DBLP:conf/emnlp/FunakiN15}
Ruka Funaki and Hideki Nakayama.
\newblock Image-mediated learning for zero-shot cross-lingual document
  retrieval.
\newblock In Llu{\'{\i}}s M{\`{a}}rquez, Chris Callison{-}Burch, Jian Su,
  Daniele Pighin, and Yuval Marton, editors, \emph{Proceedings of the 2015
  Conference on Empirical Methods in Natural Language Processing, {EMNLP} 2015,
  Lisbon, Portugal, September 17-21, 2015}, pages 585--590. The Association for
  Computational Linguistics, 2015.
\newblock \doi{10.18653/v1/d15-1070}.
\newblock URL \url{https://doi.org/10.18653/v1/d15-1070}.

\bibitem[Gangi et~al.(2019)Gangi, Cattoni, Bentivogli, Negri, and
  Turchi]{DBLP:conf/naacl/GangiCBNT19}
Mattia Antonino~Di Gangi, Roldano Cattoni, Luisa Bentivogli, Matteo Negri, and
  Marco Turchi.
\newblock Must-c: a multilingual speech translation corpus.
\newblock In Jill Burstein, Christy Doran, and Thamar Solorio, editors,
  \emph{Proceedings of the 2019 Conference of the North American Chapter of the
  Association for Computational Linguistics: Human Language Technologies,
  {NAACL-HLT} 2019, Minneapolis, MN, USA, June 2-7, 2019, Volume 1 (Long and
  Short Papers)}, pages 2012--2017. Association for Computational Linguistics,
  2019.
\newblock \doi{10.18653/v1/n19-1202}.
\newblock URL \url{https://doi.org/10.18653/v1/n19-1202}.

\bibitem[Gao et~al.(2015)Gao, Mao, Zhou, Huang, Wang, and
  Xu]{DBLP:conf/nips/GaoMZHWX15}
Haoyuan Gao, Junhua Mao, Jie Zhou, Zhiheng Huang, Lei Wang, and Wei Xu.
\newblock Are you talking to a machine? dataset and methods for multilingual
  image question.
\newblock In Corinna Cortes, Neil~D. Lawrence, Daniel~D. Lee, Masashi Sugiyama,
  and Roman Garnett, editors, \emph{Advances in Neural Information Processing
  Systems 28: Annual Conference on Neural Information Processing Systems 2015,
  December 7-12, 2015, Montreal, Quebec, Canada}, pages 2296--2304, 2015.
\newblock URL
  \url{https://proceedings.neurips.cc/paper/2015/hash/fb508ef074ee78a0e58c68be06d8a2eb-Abstract.html}.

\bibitem[Garg et~al.(2019)Garg, Peitz, Nallasamy, and
  Paulik]{DBLP:conf/emnlp/GargPNP19}
Sarthak Garg, Stephan Peitz, Udhyakumar Nallasamy, and Matthias Paulik.
\newblock Jointly learning to align and translate with transformer models.
\newblock In Kentaro Inui, Jing Jiang, Vincent Ng, and Xiaojun Wan, editors,
  \emph{Proceedings of the 2019 Conference on Empirical Methods in Natural
  Language Processing and the 9th International Joint Conference on Natural
  Language Processing, {EMNLP-IJCNLP} 2019, Hong Kong, China, November 3-7,
  2019}, pages 4452--4461. Association for Computational Linguistics, 2019.
\newblock \doi{10.18653/v1/D19-1453}.
\newblock URL \url{https://doi.org/10.18653/v1/D19-1453}.

\bibitem[Gella et~al.(2017)Gella, Sennrich, Keller, and
  Lapata]{gella-etal-2017-image}
Spandana Gella, Rico Sennrich, Frank Keller, and Mirella Lapata.
\newblock Image pivoting for learning multilingual multimodal representations.
\newblock In \emph{Proceedings of the 2017 Conference on Empirical Methods in
  Natural Language Processing}, pages 2839--2845, Copenhagen, Denmark,
  September 2017. Association for Computational Linguistics.
\newblock \doi{10.18653/v1/D17-1303}.
\newblock URL \url{https://aclanthology.org/D17-1303}.

\bibitem[Gella et~al.(2019)Gella, Elliott, and Keller]{gella-etal-2019-cross}
Spandana Gella, Desmond Elliott, and Frank Keller.
\newblock Cross-lingual visual verb sense disambiguation.
\newblock In \emph{Proceedings of the 2019 Conference of the North {A}merican
  Chapter of the Association for Computational Linguistics: Human Language
  Technologies, Volume 1 (Long and Short Papers)}, pages 1998--2004,
  Minneapolis, Minnesota, June 2019. Association for Computational Linguistics.
\newblock \doi{10.18653/v1/N19-1200}.
\newblock URL \url{https://aclanthology.org/N19-1200}.

\bibitem[Gong et~al.(2014)Gong, Ke, Isard, and
  Lazebnik]{DBLP:journals/ijcv/GongKIL14}
Yunchao Gong, Qifa Ke, Michael Isard, and Svetlana Lazebnik.
\newblock A multi-view embedding space for modeling internet images, tags, and
  their semantics.
\newblock \emph{Int. J. Comput. Vis.}, 106\penalty0 (2):\penalty0 210--233,
  2014.
\newblock \doi{10.1007/s11263-013-0658-4}.
\newblock URL \url{https://doi.org/10.1007/s11263-013-0658-4}.

\bibitem[Goyal et~al.(2017)Goyal, Khot, Summers{-}Stay, Batra, and
  Parikh]{DBLP:conf/cvpr/GoyalKSBP17}
Yash Goyal, Tejas Khot, Douglas Summers{-}Stay, Dhruv Batra, and Devi Parikh.
\newblock Making the {V} in {VQA} matter: Elevating the role of image
  understanding in visual question answering.
\newblock In \emph{2017 {IEEE} Conference on Computer Vision and Pattern
  Recognition, {CVPR} 2017, Honolulu, HI, USA, July 21-26, 2017}, pages
  6325--6334. {IEEE} Computer Society, 2017.
\newblock \doi{10.1109/CVPR.2017.670}.
\newblock URL \url{https://doi.org/10.1109/CVPR.2017.670}.

\bibitem[Graham et~al.(2020)Graham, Haddow, and
  Koehn]{DBLP:conf/emnlp/GrahamHK20}
Yvette Graham, Barry Haddow, and Philipp Koehn.
\newblock Statistical power and translationese in machine translation
  evaluation.
\newblock In Bonnie Webber, Trevor Cohn, Yulan He, and Yang Liu, editors,
  \emph{Proceedings of the 2020 Conference on Empirical Methods in Natural
  Language Processing, {EMNLP} 2020, Online, November 16-20, 2020}, pages
  72--81. Association for Computational Linguistics, 2020.
\newblock \doi{10.18653/v1/2020.emnlp-main.6}.
\newblock URL \url{https://doi.org/10.18653/v1/2020.emnlp-main.6}.

\bibitem[Gritta and Iacobacci(2021)]{DBLP:conf/acl/GrittaI21}
Milan Gritta and Ignacio Iacobacci.
\newblock Xeroalign: Zero-shot cross-lingual transformer alignment.
\newblock In Chengqing Zong, Fei Xia, Wenjie Li, and Roberto Navigli, editors,
  \emph{Findings of the Association for Computational Linguistics: {ACL/IJCNLP}
  2021, Online Event, August 1-6, 2021}, volume {ACL/IJCNLP} 2021 of
  \emph{Findings of {ACL}}, pages 371--381. Association for Computational
  Linguistics, 2021.
\newblock \doi{10.18653/v1/2021.findings-acl.32}.
\newblock URL \url{https://doi.org/10.18653/v1/2021.findings-acl.32}.

\bibitem[Gritta et~al.(2022)Gritta, Hu, and
  Iacobacci]{DBLP:conf/acl/GrittaHI22}
Milan Gritta, Ruoyu Hu, and Ignacio Iacobacci.
\newblock Crossaligner {\&} co: Zero-shot transfer methods for task-oriented
  cross-lingual natural language understanding.
\newblock In Smaranda Muresan, Preslav Nakov, and Aline Villavicencio, editors,
  \emph{Findings of the Association for Computational Linguistics: {ACL} 2022,
  Dublin, Ireland, May 22-27, 2022}, pages 4048--4061. Association for
  Computational Linguistics, 2022.
\newblock URL \url{https://aclanthology.org/2022.findings-acl.319}.

\bibitem[Grubinger et~al.(2006)Grubinger, Clough, M{\"u}ller, and
  Deselaers]{grubinger2006iapr}
Michael Grubinger, Paul Clough, Henning M{\"u}ller, and Thomas Deselaers.
\newblock The iapr tc-12 benchmark: A new evaluation resource for visual
  information systems.
\newblock In \emph{International workshop ontoImage}, volume~2, 2006.

\bibitem[Gu et~al.(2018)Gu, Joty, Cai, and Wang]{DBLP:conf/eccv/GuJCW18}
Jiuxiang Gu, Shafiq~R. Joty, Jianfei Cai, and Gang Wang.
\newblock Unpaired image captioning by language pivoting.
\newblock In Vittorio Ferrari, Martial Hebert, Cristian Sminchisescu, and Yair
  Weiss, editors, \emph{Computer Vision - {ECCV} 2018 - 15th European
  Conference, Munich, Germany, September 8-14, 2018, Proceedings, Part {I}},
  volume 11205 of \emph{Lecture Notes in Computer Science}, pages 519--535.
  Springer, 2018.
\newblock \doi{10.1007/978-3-030-01246-5\_31}.
\newblock URL \url{https://doi.org/10.1007/978-3-030-01246-5\_31}.

\bibitem[Gupta et~al.(2018)Gupta, Lenka, Ekbal, and
  Bhattacharyya]{DBLP:conf/conll/GuptaLEB18}
Deepak Gupta, Pabitra Lenka, Asif Ekbal, and Pushpak Bhattacharyya.
\newblock Uncovering code-mixed challenges: {A} framework for linguistically
  driven question generation and neural based question answering.
\newblock In Anna Korhonen and Ivan Titov, editors, \emph{Proceedings of the
  22nd Conference on Computational Natural Language Learning, CoNLL 2018,
  Brussels, Belgium, October 31 - November 1, 2018}, pages 119--130.
  Association for Computational Linguistics, 2018.
\newblock \doi{10.18653/v1/k18-1012}.
\newblock URL \url{https://doi.org/10.18653/v1/k18-1012}.

\bibitem[Gupta et~al.(2020)Gupta, Lenka, Ekbal, and
  Bhattacharyya]{gupta-etal-2020-unified}
Deepak Gupta, Pabitra Lenka, Asif Ekbal, and Pushpak Bhattacharyya.
\newblock A unified framework for multilingual and code-mixed visual question
  answering.
\newblock In \emph{Proceedings of the 1st Conference of the Asia-Pacific
  Chapter of the Association for Computational Linguistics and the 10th
  International Joint Conference on Natural Language Processing}, pages
  900--913, Suzhou, China, December 2020. Association for Computational
  Linguistics.
\newblock URL \url{https://aclanthology.org/2020.aacl-main.90}.

\bibitem[Gupta et~al.(2021)Gupta, Gautam, and Mamidi]{gupta-etal-2021-vita}
Kshitij Gupta, Devansh Gautam, and Radhika Mamidi.
\newblock {V}i{TA}: Visual-linguistic translation by aligning object tags.
\newblock In \emph{Proceedings of the 8th Workshop on Asian Translation
  (WAT2021)}, pages 166--173, Online, August 2021. Association for
  Computational Linguistics.
\newblock \doi{10.18653/v1/2021.wat-1.19}.
\newblock URL \url{https://aclanthology.org/2021.wat-1.19}.

\bibitem[Hedderich et~al.(2021)Hedderich, Lange, Adel, Str{\"{o}}tgen, and
  Klakow]{DBLP:conf/naacl/HedderichLASK21}
Michael~A. Hedderich, Lukas Lange, Heike Adel, Jannik Str{\"{o}}tgen, and
  Dietrich Klakow.
\newblock A survey on recent approaches for natural language processing in
  low-resource scenarios.
\newblock In Kristina Toutanova, Anna Rumshisky, Luke Zettlemoyer, Dilek
  Hakkani{-}T{\"{u}}r, Iz~Beltagy, Steven Bethard, Ryan Cotterell, Tanmoy
  Chakraborty, and Yichao Zhou, editors, \emph{Proceedings of the 2021
  Conference of the North American Chapter of the Association for Computational
  Linguistics: Human Language Technologies, {NAACL-HLT} 2021, Online, June
  6-11, 2021}, pages 2545--2568. Association for Computational Linguistics,
  2021.
\newblock \doi{10.18653/v1/2021.naacl-main.201}.
\newblock URL \url{https://doi.org/10.18653/v1/2021.naacl-main.201}.

\bibitem[Hitschler et~al.(2016)Hitschler, Schamoni, and
  Riezler]{DBLP:conf/acl/HitschlerSR16}
Julian Hitschler, Shigehiko Schamoni, and Stefan Riezler.
\newblock Multimodal pivots for image caption translation.
\newblock In \emph{Proceedings of the 54th Annual Meeting of the Association
  for Computational Linguistics, {ACL} 2016, August 7-12, 2016, Berlin,
  Germany, Volume 1: Long Papers}. The Association for Computer Linguistics,
  2016.
\newblock \doi{10.18653/v1/p16-1227}.
\newblock URL \url{https://doi.org/10.18653/v1/p16-1227}.

\bibitem[Huang et~al.(2019{\natexlab{a}})Huang, Liang, Duan, Gong, Shou, Jiang,
  and Zhou]{DBLP:conf/emnlp/HuangLDGSJZ19}
Haoyang Huang, Yaobo Liang, Nan Duan, Ming Gong, Linjun Shou, Daxin Jiang, and
  Ming Zhou.
\newblock Unicoder: {A} universal language encoder by pre-training with
  multiple cross-lingual tasks.
\newblock In Kentaro Inui, Jing Jiang, Vincent Ng, and Xiaojun Wan, editors,
  \emph{Proceedings of the 2019 Conference on Empirical Methods in Natural
  Language Processing and the 9th International Joint Conference on Natural
  Language Processing, {EMNLP-IJCNLP} 2019, Hong Kong, China, November 3-7,
  2019}, pages 2485--2494. Association for Computational Linguistics,
  2019{\natexlab{a}}.
\newblock \doi{10.18653/v1/D19-1252}.
\newblock URL \url{https://doi.org/10.18653/v1/D19-1252}.

\bibitem[Huang et~al.(2019{\natexlab{b}})Huang, Chang, and
  Hauptmann]{huang-etal-2019-multi}
Po-Yao Huang, Xiaojun Chang, and Alexander Hauptmann.
\newblock Multi-head attention with diversity for learning grounded
  multilingual multimodal representations.
\newblock In \emph{Proceedings of the 2019 Conference on Empirical Methods in
  Natural Language Processing and the 9th International Joint Conference on
  Natural Language Processing (EMNLP-IJCNLP)}, pages 1461--1467, Hong Kong,
  China, November 2019{\natexlab{b}}. Association for Computational
  Linguistics.
\newblock \doi{10.18653/v1/D19-1154}.
\newblock URL \url{https://aclanthology.org/D19-1154}.

\bibitem[Huang et~al.(2020)Huang, Hu, Chang, and
  Hauptmann]{huang-etal-2020-unsupervised-multimodal}
Po-Yao Huang, Junjie Hu, Xiaojun Chang, and Alexander Hauptmann.
\newblock Unsupervised multimodal neural machine translation with pseudo visual
  pivoting.
\newblock In \emph{Proceedings of the 58th Annual Meeting of the Association
  for Computational Linguistics}, pages 8226--8237, Online, July 2020.
  Association for Computational Linguistics.
\newblock \doi{10.18653/v1/2020.acl-main.731}.
\newblock URL \url{https://aclanthology.org/2020.acl-main.731}.

\bibitem[Huang et~al.(2021)Huang, Patrick, Hu, Neubig, Metze, and
  Hauptmann]{huang-etal-2021-multilingual}
Po-Yao Huang, Mandela Patrick, Junjie Hu, Graham Neubig, Florian Metze, and
  Alexander Hauptmann.
\newblock Multilingual multimodal pre-training for zero-shot cross-lingual
  transfer of vision-language models.
\newblock In \emph{Proceedings of the 2021 Conference of the North American
  Chapter of the Association for Computational Linguistics: Human Language
  Technologies}, pages 2443--2459, Online, June 2021. Association for
  Computational Linguistics.
\newblock \doi{10.18653/v1/2021.naacl-main.195}.
\newblock URL \url{https://aclanthology.org/2021.naacl-main.195}.

\bibitem[Imankulova et~al.(2020)Imankulova, Kaneko, Hirasawa, and
  Komachi]{imankulova-etal-2020-towards}
Aizhan Imankulova, Masahiro Kaneko, Tosho Hirasawa, and Mamoru Komachi.
\newblock Towards multimodal simultaneous neural machine translation.
\newblock In \emph{Proceedings of the Fifth Conference on Machine Translation},
  pages 594--603, Online, November 2020. Association for Computational
  Linguistics.
\newblock URL \url{https://aclanthology.org/2020.wmt-1.70}.

\bibitem[Jones and Muftic(2020)]{jones-muftic-2020-endangered}
Kerry Jones and Sanjin Muftic.
\newblock Endangered {A}frican languages featured in a digital collection: The
  case of the {K}homani {S}an, {H}ugh {B}rody {C}ollection.
\newblock In \emph{Proceedings of the first workshop on Resources for African
  Indigenous Languages}, pages 1--8, Marseille, France, May 2020. European
  Language Resources Association (ELRA).
\newblock ISBN 979-10-95546-60-3.
\newblock URL \url{https://aclanthology.org/2020.rail-1.1}.

\bibitem[K{\'a}d{\'a}r et~al.(2018)K{\'a}d{\'a}r, Elliott, C{\^o}t{\'e},
  Chrupa{\l}a, and Alishahi]{kadar-etal-2018-lessons}
{\'A}kos K{\'a}d{\'a}r, Desmond Elliott, Marc-Alexandre C{\^o}t{\'e}, Grzegorz
  Chrupa{\l}a, and Afra Alishahi.
\newblock Lessons learned in multilingual grounded language learning.
\newblock In \emph{Proceedings of the 22nd Conference on Computational Natural
  Language Learning}, pages 402--412, Brussels, Belgium, October 2018.
  Association for Computational Linguistics.
\newblock \doi{10.18653/v1/K18-1039}.
\newblock URL \url{https://aclanthology.org/K18-1039}.

\bibitem[Karakanta et~al.(2020)Karakanta, Negri, and
  Turchi]{karakanta-etal-2020-must}
Alina Karakanta, Matteo Negri, and Marco Turchi.
\newblock {M}u{ST}-cinema: a speech-to-subtitles corpus.
\newblock In \emph{Proceedings of the 12th Language Resources and Evaluation
  Conference}, pages 3727--3734, Marseille, France, May 2020. European Language
  Resources Association.
\newblock ISBN 979-10-95546-34-4.
\newblock URL \url{https://aclanthology.org/2020.lrec-1.460}.

\bibitem[Karakanta et~al.(2021)Karakanta, Papi, Negri, and
  Turchi]{karakanta-etal-2021-simultaneous}
Alina Karakanta, Sara Papi, Matteo Negri, and Marco Turchi.
\newblock Simultaneous speech translation for live subtitling: from delay to
  display.
\newblock In \emph{Proceedings of the 1st Workshop on Automatic Spoken Language
  Translation in Real-World Settings (ASLTRW)}, pages 35--48, Virtual, August
  2021. Association for Machine Translation in the Americas.
\newblock URL \url{https://aclanthology.org/2021.mtsummit-asltrw.4}.

\bibitem[Kay et~al.(2017)Kay, Carreira, Simonyan, Zhang, Hillier,
  Vijayanarasimhan, Viola, Green, Back, Natsev, Suleyman, and
  Zisserman]{DBLP:journals/corr/KayCSZHVVGBNSZ17}
Will Kay, Jo{\~{a}}o Carreira, Karen Simonyan, Brian Zhang, Chloe Hillier,
  Sudheendra Vijayanarasimhan, Fabio Viola, Tim Green, Trevor Back, Paul
  Natsev, Mustafa Suleyman, and Andrew Zisserman.
\newblock The kinetics human action video dataset.
\newblock \emph{CoRR}, abs/1705.06950, 2017.
\newblock URL \url{http://arxiv.org/abs/1705.06950}.

\bibitem[Kiros et~al.(2014)Kiros, Salakhutdinov, and
  Zemel]{DBLP:journals/corr/KirosSZ14}
Ryan Kiros, Ruslan Salakhutdinov, and Richard~S. Zemel.
\newblock Unifying visual-semantic embeddings with multimodal neural language
  models.
\newblock \emph{CoRR}, abs/1411.2539, 2014.
\newblock URL \url{http://arxiv.org/abs/1411.2539}.

\bibitem[Kitaev and Klein(2018)]{DBLP:conf/acl/KleinK18}
Nikita Kitaev and Dan Klein.
\newblock Constituency parsing with a self-attentive encoder.
\newblock In Iryna Gurevych and Yusuke Miyao, editors, \emph{Proceedings of the
  56th Annual Meeting of the Association for Computational Linguistics, {ACL}
  2018, Melbourne, Australia, July 15-20, 2018, Volume 1: Long Papers}, pages
  2676--2686. Association for Computational Linguistics, 2018.
\newblock \doi{10.18653/v1/P18-1249}.
\newblock URL \url{https://aclanthology.org/P18-1249/}.

\bibitem[Koehn et~al.(2003)Koehn, Och, and Marcu]{koehn2003statistical}
Philipp Koehn, Franz~J Och, and Daniel Marcu.
\newblock Statistical phrase-based translation.
\newblock Technical report, University of Southern California Marina Del Rey
  Information Sciences Inst, 2003.

\bibitem[Koehn et~al.(2007)Koehn, Hoang, Birch, Callison{-}Burch, Federico,
  Bertoldi, Cowan, Shen, Moran, Zens, Dyer, Bojar, Constantin, and
  Herbst]{DBLP:conf/acl/KoehnHBCFBCSMZDBCH07}
Philipp Koehn, Hieu Hoang, Alexandra Birch, Chris Callison{-}Burch, Marcello
  Federico, Nicola Bertoldi, Brooke Cowan, Wade Shen, Christine Moran, Richard
  Zens, Chris Dyer, Ondrej Bojar, Alexandra Constantin, and Evan Herbst.
\newblock Moses: Open source toolkit for statistical machine translation.
\newblock In John~A. Carroll, Antal van~den Bosch, and Annie Zaenen, editors,
  \emph{{ACL} 2007, Proceedings of the 45th Annual Meeting of the Association
  for Computational Linguistics, June 23-30, 2007, Prague, Czech Republic}. The
  Association for Computational Linguistics, 2007.
\newblock URL \url{https://aclanthology.org/P07-2045/}.

\bibitem[Koeva(2021)]{koeva-2021-multilingual}
Svetla Koeva.
\newblock Multilingual image corpus: Annotation protocol.
\newblock In \emph{Proceedings of the International Conference on Recent
  Advances in Natural Language Processing (RANLP 2021)}, pages 701--707, Held
  Online, September 2021. INCOMA Ltd.
\newblock URL \url{https://aclanthology.org/2021.ranlp-1.80}.

\bibitem[Koppel and Ordan(2011)]{DBLP:conf/acl/KoppelO11}
Moshe Koppel and Noam Ordan.
\newblock Translationese and its dialects.
\newblock In Dekang Lin, Yuji Matsumoto, and Rada Mihalcea, editors, \emph{The
  49th Annual Meeting of the Association for Computational Linguistics: Human
  Language Technologies, Proceedings of the Conference, 19-24 June, 2011,
  Portland, Oregon, {USA}}, pages 1318--1326. The Association for Computer
  Linguistics, 2011.
\newblock URL \url{https://aclanthology.org/P11-1132/}.

\bibitem[Krishna et~al.(2017)Krishna, Zhu, Groth, Johnson, Hata, Kravitz, Chen,
  Kalantidis, Li, Shamma, Bernstein, and
  Fei{-}Fei]{DBLP:journals/ijcv/KrishnaZGJHKCKL17}
Ranjay Krishna, Yuke Zhu, Oliver Groth, Justin Johnson, Kenji Hata, Joshua
  Kravitz, Stephanie Chen, Yannis Kalantidis, Li{-}Jia Li, David~A. Shamma,
  Michael~S. Bernstein, and Li~Fei{-}Fei.
\newblock Visual genome: Connecting language and vision using crowdsourced
  dense image annotations.
\newblock \emph{Int. J. Comput. Vis.}, 123\penalty0 (1):\penalty0 32--73, 2017.
\newblock \doi{10.1007/s11263-016-0981-7}.
\newblock URL \url{https://doi.org/10.1007/s11263-016-0981-7}.

\bibitem[Krishnan et~al.(2021)Krishnan, Anastasopoulos, Purohit, and
  Rangwala]{DBLP:journals/corr/abs-2103-07792}
Jitin Krishnan, Antonios Anastasopoulos, Hemant Purohit, and Huzefa Rangwala.
\newblock Multilingual code-switching for zero-shot cross-lingual intent
  prediction and slot filling.
\newblock \emph{CoRR}, abs/2103.07792, 2021.
\newblock URL \url{https://arxiv.org/abs/2103.07792}.

\bibitem[Ku et~al.(2020)Ku, Anderson, Patel, Ie, and
  Baldridge]{ku-etal-2020-room}
Alexander Ku, Peter Anderson, Roma Patel, Eugene Ie, and Jason Baldridge.
\newblock Room-across-room: Multilingual vision-and-language navigation with
  dense spatiotemporal grounding.
\newblock In \emph{Proceedings of the 2020 Conference on Empirical Methods in
  Natural Language Processing (EMNLP)}, pages 4392--4412, Online, November
  2020. Association for Computational Linguistics.
\newblock \doi{10.18653/v1/2020.emnlp-main.356}.
\newblock URL \url{https://aclanthology.org/2020.emnlp-main.356}.

\bibitem[Lala and Specia(2018)]{DBLP:conf/lrec/LalaS18}
Chiraag Lala and Lucia Specia.
\newblock Multimodal lexical translation.
\newblock In Nicoletta Calzolari, Khalid Choukri, Christopher Cieri, Thierry
  Declerck, Sara Goggi, K{\^{o}}iti Hasida, Hitoshi Isahara, Bente Maegaard,
  Joseph Mariani, H{\'{e}}l{\`{e}}ne Mazo, Asunci{\'{o}}n Moreno, Jan Odijk,
  Stelios Piperidis, and Takenobu Tokunaga, editors, \emph{Proceedings of the
  Eleventh International Conference on Language Resources and Evaluation,
  {LREC} 2018, Miyazaki, Japan, May 7-12, 2018}. European Language Resources
  Association {(ELRA)}, 2018.
\newblock URL
  \url{http://www.lrec-conf.org/proceedings/lrec2018/summaries/629.html}.

\bibitem[Lala et~al.(2018)Lala, Madhyastha, Scarton, and
  Specia]{lala-etal-2018-sheffield}
Chiraag Lala, Pranava~Swaroop Madhyastha, Carolina Scarton, and Lucia Specia.
\newblock {S}heffield submissions for {WMT}18 multimodal translation shared
  task.
\newblock In \emph{Proceedings of the Third Conference on Machine Translation:
  Shared Task Papers}, pages 624--631, Belgium, Brussels, October 2018.
  Association for Computational Linguistics.
\newblock \doi{10.18653/v1/W18-6442}.
\newblock URL \url{https://aclanthology.org/W18-6442}.

\bibitem[Lan et~al.(2017)Lan, Li, and Dong]{DBLP:conf/mm/LanLD17}
Weiyu Lan, Xirong Li, and Jianfeng Dong.
\newblock Fluency-guided cross-lingual image captioning.
\newblock In Qiong Liu, Rainer Lienhart, Haohong Wang, Sheng{-}Wei~"Kuan{-}Ta"
  Chen, Susanne Boll, Yi{-}Ping~Phoebe Chen, Gerald Friedland, Jia Li, and
  Shuicheng Yan, editors, \emph{Proceedings of the 2017 {ACM} on Multimedia
  Conference, {MM} 2017, Mountain View, CA, USA, October 23-27, 2017}, pages
  1549--1557. {ACM}, 2017.
\newblock \doi{10.1145/3123266.3123366}.
\newblock URL \url{https://doi.org/10.1145/3123266.3123366}.

\bibitem[Lauscher et~al.(2020)Lauscher, Ravishankar, Vulic, and
  Glavas]{DBLP:conf/emnlp/LauscherRVG20}
Anne Lauscher, Vinit Ravishankar, Ivan Vulic, and Goran Glavas.
\newblock From zero to hero: On the limitations of zero-shot language transfer
  with multilingual transformers.
\newblock In Bonnie Webber, Trevor Cohn, Yulan He, and Yang Liu, editors,
  \emph{Proceedings of the 2020 Conference on Empirical Methods in Natural
  Language Processing, {EMNLP} 2020, Online, November 16-20, 2020}, pages
  4483--4499. Association for Computational Linguistics, 2020.
\newblock \doi{10.18653/v1/2020.emnlp-main.363}.
\newblock URL \url{https://doi.org/10.18653/v1/2020.emnlp-main.363}.

\bibitem[Leviant and Reichart(2015)]{leviant2015separated}
Ira Leviant and Roi Reichart.
\newblock Separated by an un-common language: Towards judgment language
  informed vector space modeling.
\newblock \emph{arXiv preprint arXiv:1508.00106}, 2015.

\bibitem[Li et~al.(2020)Li, Duan, Fang, Gong, and
  Jiang]{DBLP:conf/aaai/LiDFGJ20}
Gen Li, Nan Duan, Yuejian Fang, Ming Gong, and Daxin Jiang.
\newblock Unicoder-vl: {A} universal encoder for vision and language by
  cross-modal pre-training.
\newblock In \emph{The Thirty-Fourth {AAAI} Conference on Artificial
  Intelligence, {AAAI} 2020, The Thirty-Second Innovative Applications of
  Artificial Intelligence Conference, {IAAI} 2020, The Tenth {AAAI} Symposium
  on Educational Advances in Artificial Intelligence, {EAAI} 2020, New York,
  NY, USA, February 7-12, 2020}, pages 11336--11344. {AAAI} Press, 2020.
\newblock URL \url{https://ojs.aaai.org/index.php/AAAI/article/view/6795}.

\bibitem[Li et~al.(2019{\natexlab{a}})Li, Yatskar, Yin, Hsieh, and
  Chang]{DBLP:journals/corr/abs-1908-03557}
Liunian~Harold Li, Mark Yatskar, Da~Yin, Cho{-}Jui Hsieh, and Kai{-}Wei Chang.
\newblock Visualbert: {A} simple and performant baseline for vision and
  language.
\newblock \emph{CoRR}, abs/1908.03557, 2019{\natexlab{a}}.
\newblock URL \url{http://arxiv.org/abs/1908.03557}.

\bibitem[Li et~al.(2016)Li, Lan, Dong, and Liu]{DBLP:conf/mir/LiLDL16}
Xirong Li, Weiyu Lan, Jianfeng Dong, and Hailong Liu.
\newblock Adding chinese captions to images.
\newblock In John~R. Kender, John~R. Smith, Jiebo Luo, Susanne Boll, and
  Winston~H. Hsu, editors, \emph{Proceedings of the 2016 {ACM} on International
  Conference on Multimedia Retrieval, {ICMR} 2016, New York, New York, USA,
  June 6-9, 2016}, pages 271--275. {ACM}, 2016.
\newblock \doi{10.1145/2911996.2912049}.
\newblock URL \url{https://doi.org/10.1145/2911996.2912049}.

\bibitem[Li et~al.(2019{\natexlab{b}})Li, Xu, Wang, Lan, Jia, Yang, and
  Xu]{DBLP:journals/tmm/LiXWLJYX19}
Xirong Li, Chaoxi Xu, Xiaoxu Wang, Weiyu Lan, Zhengxiong Jia, Gang Yang, and
  Jieping Xu.
\newblock {COCO-CN} for cross-lingual image tagging, captioning, and retrieval.
\newblock \emph{{IEEE} Trans. Multim.}, 21\penalty0 (9):\penalty0 2347--2360,
  2019{\natexlab{b}}.
\newblock \doi{10.1109/TMM.2019.2896494}.
\newblock URL \url{https://doi.org/10.1109/TMM.2019.2896494}.

\bibitem[Lin et~al.(2014)Lin, Maire, Belongie, Hays, Perona, Ramanan,
  Doll{\'{a}}r, and Zitnick]{DBLP:conf/eccv/LinMBHPRDZ14}
Tsung{-}Yi Lin, Michael Maire, Serge~J. Belongie, James Hays, Pietro Perona,
  Deva Ramanan, Piotr Doll{\'{a}}r, and C.~Lawrence Zitnick.
\newblock Microsoft {COCO:} common objects in context.
\newblock In David~J. Fleet, Tom{\'{a}}s Pajdla, Bernt Schiele, and Tinne
  Tuytelaars, editors, \emph{Computer Vision - {ECCV} 2014 - 13th European
  Conference, Zurich, Switzerland, September 6-12, 2014, Proceedings, Part
  {V}}, volume 8693 of \emph{Lecture Notes in Computer Science}, pages
  740--755. Springer, 2014.
\newblock \doi{10.1007/978-3-319-10602-1\_48}.
\newblock URL \url{https://doi.org/10.1007/978-3-319-10602-1\_48}.

\bibitem[Liu et~al.(2021)Liu, Bugliarello, Ponti, Reddy, Collier, and
  Elliott]{liu-etal-2021-visually}
Fangyu Liu, Emanuele Bugliarello, Edoardo~Maria Ponti, Siva Reddy, Nigel
  Collier, and Desmond Elliott.
\newblock Visually grounded reasoning across languages and cultures.
\newblock In \emph{Proceedings of the 2021 Conference on Empirical Methods in
  Natural Language Processing}, pages 10467--10485, Online and Punta Cana,
  Dominican Republic, November 2021. Association for Computational Linguistics.
\newblock \doi{10.18653/v1/2021.emnlp-main.818}.
\newblock URL \url{https://aclanthology.org/2021.emnlp-main.818}.

\bibitem[Liu et~al.(2020)Liu, Gu, Goyal, Li, Edunov, Ghazvininejad, Lewis, and
  Zettlemoyer]{DBLP:journals/tacl/LiuGGLEGLZ20}
Yinhan Liu, Jiatao Gu, Naman Goyal, Xian Li, Sergey Edunov, Marjan
  Ghazvininejad, Mike Lewis, and Luke Zettlemoyer.
\newblock Multilingual denoising pre-training for neural machine translation.
\newblock \emph{Trans. Assoc. Comput. Linguistics}, 8:\penalty0 726--742, 2020.
\newblock URL \url{https://transacl.org/ojs/index.php/tacl/article/view/2107}.

\bibitem[Lu et~al.(2019)Lu, Batra, Parikh, and Lee]{DBLP:conf/nips/LuBPL19}
Jiasen Lu, Dhruv Batra, Devi Parikh, and Stefan Lee.
\newblock Vilbert: Pretraining task-agnostic visiolinguistic representations
  for vision-and-language tasks.
\newblock In Hanna~M. Wallach, Hugo Larochelle, Alina Beygelzimer, Florence
  d'Alch{\'{e}}{-}Buc, Emily~B. Fox, and Roman Garnett, editors, \emph{Advances
  in Neural Information Processing Systems 32: Annual Conference on Neural
  Information Processing Systems 2019, NeurIPS 2019, December 8-14, 2019,
  Vancouver, BC, Canada}, pages 13--23, 2019.
\newblock URL
  \url{https://proceedings.neurips.cc/paper/2019/hash/c74d97b01eae257e44aa9d5bade97baf-Abstract.html}.

\bibitem[Ma et~al.(2019)Ma, Huang, Xiong, Zheng, Liu, Zheng, Zhang, He, Liu,
  Li, Wu, and Wang]{DBLP:conf/acl/MaHXZLZZHLLWW19}
Mingbo Ma, Liang Huang, Hao Xiong, Renjie Zheng, Kaibo Liu, Baigong Zheng,
  Chuanqiang Zhang, Zhongjun He, Hairong Liu, Xing Li, Hua Wu, and Haifeng
  Wang.
\newblock {STACL:} simultaneous translation with implicit anticipation and
  controllable latency using prefix-to-prefix framework.
\newblock In Anna Korhonen, David~R. Traum, and Llu{\'{\i}}s M{\`{a}}rquez,
  editors, \emph{Proceedings of the 57th Conference of the Association for
  Computational Linguistics, {ACL} 2019, Florence, Italy, July 28- August 2,
  2019, Volume 1: Long Papers}, pages 3025--3036. Association for Computational
  Linguistics, 2019.
\newblock \doi{10.18653/v1/p19-1289}.
\newblock URL \url{https://doi.org/10.18653/v1/p19-1289}.

\bibitem[Mitzalis et~al.(2021)Mitzalis, Caglayan, Madhyastha, and
  Specia]{mitzalis-etal-2021-bertgen}
Faidon Mitzalis, Ozan Caglayan, Pranava Madhyastha, and Lucia Specia.
\newblock {BERTG}en: Multi-task generation through {BERT}.
\newblock In \emph{Proceedings of the 59th Annual Meeting of the Association
  for Computational Linguistics and the 11th International Joint Conference on
  Natural Language Processing (Volume 1: Long Papers)}, pages 6440--6455,
  Online, August 2021. Association for Computational Linguistics.
\newblock \doi{10.18653/v1/2021.acl-long.503}.
\newblock URL \url{https://aclanthology.org/2021.acl-long.503}.

\bibitem[Mohammadshahi et~al.(2019)Mohammadshahi, Lebret, and
  Aberer]{mohammadshahi-etal-2019-aligning-multilingual}
Alireza Mohammadshahi, R{\'e}mi Lebret, and Karl Aberer.
\newblock Aligning multilingual word embeddings for cross-modal retrieval task.
\newblock In \emph{Proceedings of the Second Workshop on Fact Extraction and
  VERification (FEVER)}, pages 27--33, Hong Kong, China, November 2019.
  Association for Computational Linguistics.
\newblock \doi{10.18653/v1/D19-6605}.
\newblock URL \url{https://aclanthology.org/D19-6605}.

\bibitem[Moneglia and Varvara(2020)]{moneglia-varvara-2020-annotation}
Massimo Moneglia and Rossella Varvara.
\newblock The annotation of thematic structure and alternations face to the
  semantic variation of action verbs. current trends in the {IMAGACT} ontology.
\newblock In \emph{16th Joint ACL - ISO Workshop on Interoperable Semantic
  Annotation PROCEEDINGS}, pages 67--74, Marseille, May 2020. European Language
  Resources Association.
\newblock ISBN 979-10-95546-48-1.
\newblock URL \url{https://aclanthology.org/2020.isa-1.8}.

\bibitem[Moneglia et~al.(2014{\natexlab{a}})Moneglia, Brown, Frontini,
  Gagliardi, Khan, Monachini, and Panunzi]{DBLP:conf/lrec/MonegliaBFGKMP14}
Massimo Moneglia, Susan Brown, Francesca Frontini, Gloria Gagliardi, Anas~Fahad
  Khan, Monica Monachini, and Alessandro Panunzi.
\newblock The {IMAGACT} visual ontology. an extendable multilingual
  infrastructure for the representation of lexical encoding of action.
\newblock In Nicoletta Calzolari, Khalid Choukri, Thierry Declerck, Hrafn
  Loftsson, Bente Maegaard, Joseph Mariani, Asunci{\'{o}}n Moreno, Jan Odijk,
  and Stelios Piperidis, editors, \emph{Proceedings of the Ninth International
  Conference on Language Resources and Evaluation, {LREC} 2014, Reykjavik,
  Iceland, May 26-31, 2014}, pages 3425--3432. European Language Resources
  Association {(ELRA)}, 2014{\natexlab{a}}.
\newblock URL
  \url{http://www.lrec-conf.org/proceedings/lrec2014/summaries/318.html}.

\bibitem[Moneglia et~al.(2014{\natexlab{b}})Moneglia, Brown, Frontini,
  Gagliardi, Khan, Monachini, and Panunzi]{moneglia-etal-2014-imagact}
Massimo Moneglia, Susan Brown, Francesca Frontini, Gloria Gagliardi, Fahad
  Khan, Monica Monachini, and Alessandro Panunzi.
\newblock The {IMAGACT} visual ontology. an extendable multilingual
  infrastructure for the representation of lexical encoding of action.
\newblock In \emph{Proceedings of the Ninth International Conference on
  Language Resources and Evaluation ({LREC}'14)}, pages 3425--3432, Reykjavik,
  Iceland, May 2014{\natexlab{b}}. European Language Resources Association
  (ELRA).
\newblock URL
  \url{http://www.lrec-conf.org/proceedings/lrec2014/pdf/318_Paper.pdf}.

\bibitem[Myers-Scotton(1997)]{myers1997duelling}
Carol Myers-Scotton.
\newblock \emph{Duelling languages: Grammatical structure in codeswitching}.
\newblock Oxford University Press, 1997.

\bibitem[Nakayama et~al.(2020{\natexlab{a}})Nakayama, Tamura, and
  Ninomiya]{nakayama-etal-2020-visually}
Hideki Nakayama, Akihiro Tamura, and Takashi Ninomiya.
\newblock A visually-grounded parallel corpus with phrase-to-region linking.
\newblock In \emph{Proceedings of the 12th Language Resources and Evaluation
  Conference}, pages 4204--4210, Marseille, France, May 2020{\natexlab{a}}.
  European Language Resources Association.
\newblock ISBN 979-10-95546-34-4.
\newblock URL \url{https://aclanthology.org/2020.lrec-1.518}.

\bibitem[Nakayama et~al.(2020{\natexlab{b}})Nakayama, Tamura, and
  Ninomiya]{nakayama2020visually}
Hideki Nakayama, Akihiro Tamura, and Takashi Ninomiya.
\newblock A visually-grounded parallel corpus with phrase-to-region linking.
\newblock In \emph{Proceedings of The 12th Language Resources and Evaluation
  Conference}, pages 4204--4210, 2020{\natexlab{b}}.

\bibitem[Ni et~al.(2021)Ni, Huang, Su, Cui, Bharti, Wang, Zhang, and
  Duan]{DBLP:conf/cvpr/NiHSCBW0D21}
Minheng Ni, Haoyang Huang, Lin Su, Edward Cui, Taroon Bharti, Lijuan Wang,
  Dongdong Zhang, and Nan Duan.
\newblock {M3P:} learning universal representations via multitask multilingual
  multimodal pre-training.
\newblock In \emph{{IEEE} Conference on Computer Vision and Pattern
  Recognition, {CVPR} 2021, virtual, June 19-25, 2021}, pages 3977--3986.
  Computer Vision Foundation / {IEEE}, 2021.
\newblock URL
  \url{https://openaccess.thecvf.com/content/CVPR2021/html/Ni\_M3P\_Learning\_Universal\_Representations\_via\_Multitask\_Multilingual\_Multimodal\_Pre-Training\_CVPR\_2021\_paper.html}.

\bibitem[Nishihara et~al.(2020)Nishihara, Tamura, Ninomiya, Omote, and
  Nakayama]{nishihara-etal-2020-supervised}
Tetsuro Nishihara, Akihiro Tamura, Takashi Ninomiya, Yutaro Omote, and Hideki
  Nakayama.
\newblock Supervised visual attention for multimodal neural machine
  translation.
\newblock In \emph{Proceedings of the 28th International Conference on
  Computational Linguistics}, pages 4304--4314, Barcelona, Spain (Online),
  December 2020. International Committee on Computational Linguistics.
\newblock \doi{10.18653/v1/2020.coling-main.380}.
\newblock URL \url{https://aclanthology.org/2020.coling-main.380}.

\bibitem[Parida and Bojar(2018)]{parida2018translating}
Shantipriya Parida and Ond{\v{r}}ej Bojar.
\newblock Translating short segments with nmt: A case study in
  english-to-hindi.
\newblock 2018.

\bibitem[Parida et~al.(2020)Parida, Motlicek, Dash, Dash, Mallick, Biswal,
  Pattnaik, Nayak, and Bojar]{parida-etal-2020-odianlps}
Shantipriya Parida, Petr Motlicek, Amulya~Ratna Dash, Satya~Ranjan Dash,
  Debasish~Kumar Mallick, Satya~Prakash Biswal, Priyanka Pattnaik,
  Biranchi~Narayan Nayak, and Ond{\v{r}}ej Bojar.
\newblock {ODIANLP}{'}s participation in {WAT}2020.
\newblock In \emph{Proceedings of the 7th Workshop on Asian Translation}, pages
  103--108, Suzhou, China, December 2020. Association for Computational
  Linguistics.
\newblock URL \url{https://aclanthology.org/2020.wat-1.10}.

\bibitem[Parida et~al.(2021)Parida, Panda, Kotwal, Dash, Dash, Sharma,
  Motlicek, and Bojar]{parida-etal-2021-nlphuts}
Shantipriya Parida, Subhadarshi Panda, Ketan Kotwal, Amulya~Ratna Dash,
  Satya~Ranjan Dash, Yashvardhan Sharma, Petr Motlicek, and Ond{\v{r}}ej Bojar.
\newblock {NLPH}ut{'}s participation at {WAT}2021.
\newblock In \emph{Proceedings of the 8th Workshop on Asian Translation
  (WAT2021)}, pages 146--154, Online, August 2021. Association for
  Computational Linguistics.
\newblock \doi{10.18653/v1/2021.wat-1.16}.
\newblock URL \url{https://aclanthology.org/2021.wat-1.16}.

\bibitem[Pfeiffer et~al.(2020)Pfeiffer, Vulic, Gurevych, and
  Ruder]{DBLP:conf/emnlp/PfeifferVGR20}
Jonas Pfeiffer, Ivan Vulic, Iryna Gurevych, and Sebastian Ruder.
\newblock {MAD-X:} an adapter-based framework for multi-task cross-lingual
  transfer.
\newblock In Bonnie Webber, Trevor Cohn, Yulan He, and Yang Liu, editors,
  \emph{Proceedings of the 2020 Conference on Empirical Methods in Natural
  Language Processing, {EMNLP} 2020, Online, November 16-20, 2020}, pages
  7654--7673. Association for Computational Linguistics, 2020.
\newblock \doi{10.18653/v1/2020.emnlp-main.617}.
\newblock URL \url{https://doi.org/10.18653/v1/2020.emnlp-main.617}.

\bibitem[Pfeiffer et~al.(2022)Pfeiffer, Geigle, Kamath, Steitz, Roth, Vulic,
  and Gurevych]{DBLP:conf/acl/PfeifferGKS0VG22}
Jonas Pfeiffer, Gregor Geigle, Aishwarya Kamath, Jan{-}Martin~O. Steitz, Stefan
  Roth, Ivan Vulic, and Iryna Gurevych.
\newblock xgqa: Cross-lingual visual question answering.
\newblock In Smaranda Muresan, Preslav Nakov, and Aline Villavicencio, editors,
  \emph{Findings of the Association for Computational Linguistics: {ACL} 2022,
  Dublin, Ireland, May 22-27, 2022}, pages 2497--2511. Association for
  Computational Linguistics, 2022.
\newblock \doi{10.18653/v1/2022.findings-acl.196}.
\newblock URL \url{https://doi.org/10.18653/v1/2022.findings-acl.196}.

\bibitem[Plummer et~al.(2015)Plummer, Wang, Cervantes, Caicedo, Hockenmaier,
  and Lazebnik]{plummer2015flickr30k}
Bryan~A Plummer, Liwei Wang, Chris~M Cervantes, Juan~C Caicedo, Julia
  Hockenmaier, and Svetlana Lazebnik.
\newblock Flickr30k entities: Collecting region-to-phrase correspondences for
  richer image-to-sentence models.
\newblock In \emph{Proceedings of the IEEE international conference on computer
  vision}, pages 2641--2649, 2015.

\bibitem[Poignant et~al.(2016)Poignant, Budnik, Bredin, Barras, Stefas,
  Bruneau, Adda, Besacier, Ekenel, Francopoulo, Hernando, Mariani, Morros,
  Qu{\'e}not, Rosset, and Tamisier]{poignant-etal-2016-camomile}
Johann Poignant, Mateusz Budnik, Herv{\'e} Bredin, Claude Barras, Mickael
  Stefas, Pierrick Bruneau, Gilles Adda, Laurent Besacier, Hazim Ekenel, Gil
  Francopoulo, Javier Hernando, Joseph Mariani, Ramon Morros, Georges
  Qu{\'e}not, Sophie Rosset, and Thomas Tamisier.
\newblock The {CAMOMILE} collaborative annotation platform for multi-modal,
  multi-lingual and multi-media documents.
\newblock In \emph{Proceedings of the Tenth International Conference on
  Language Resources and Evaluation ({LREC}'16)}, pages 1421--1425,
  Portoro{\v{z}}, Slovenia, May 2016. European Language Resources Association
  (ELRA).
\newblock URL \url{https://aclanthology.org/L16-1226}.

\bibitem[Qin et~al.(2020)Qin, Ni, Zhang, and Che]{DBLP:conf/ijcai/QinN0C20}
Libo Qin, Minheng Ni, Yue Zhang, and Wanxiang Che.
\newblock Cosda-ml: Multi-lingual code-switching data augmentation for
  zero-shot cross-lingual {NLP}.
\newblock In Christian Bessiere, editor, \emph{Proceedings of the Twenty-Ninth
  International Joint Conference on Artificial Intelligence, {IJCAI} 2020},
  pages 3853--3860. ijcai.org, 2020.
\newblock \doi{10.24963/ijcai.2020/533}.
\newblock URL \url{https://doi.org/10.24963/ijcai.2020/533}.

\bibitem[Radford et~al.(2021)Radford, Kim, Hallacy, Ramesh, Goh, Agarwal,
  Sastry, Askell, Mishkin, Clark, Krueger, and
  Sutskever]{DBLP:conf/icml/RadfordKHRGASAM21}
Alec Radford, Jong~Wook Kim, Chris Hallacy, Aditya Ramesh, Gabriel Goh,
  Sandhini Agarwal, Girish Sastry, Amanda Askell, Pamela Mishkin, Jack Clark,
  Gretchen Krueger, and Ilya Sutskever.
\newblock Learning transferable visual models from natural language
  supervision.
\newblock In Marina Meila and Tong Zhang, editors, \emph{Proceedings of the
  38th International Conference on Machine Learning, {ICML} 2021, 18-24 July
  2021, Virtual Event}, volume 139 of \emph{Proceedings of Machine Learning
  Research}, pages 8748--8763. {PMLR}, 2021.
\newblock URL \url{http://proceedings.mlr.press/v139/radford21a.html}.

\bibitem[Raj~Khan et~al.(2021)Raj~Khan, Gupta, and
  Ekbal]{raj-khan-etal-2021-towards-developing}
Humair Raj~Khan, Deepak Gupta, and Asif Ekbal.
\newblock Towards developing a multilingual and code-mixed visual question
  answering system by knowledge distillation.
\newblock In \emph{Findings of the Association for Computational Linguistics:
  EMNLP 2021}, pages 1753--1767, Punta Cana, Dominican Republic, November 2021.
  Association for Computational Linguistics.
\newblock \doi{10.18653/v1/2021.findings-emnlp.151}.
\newblock URL \url{https://aclanthology.org/2021.findings-emnlp.151}.

\bibitem[Rajendran et~al.(2016)Rajendran, Khapra, Chandar, and
  Ravindran]{DBLP:conf/naacl/RajendranKCR16}
Janarthanan Rajendran, Mitesh~M. Khapra, Sarath Chandar, and Balaraman
  Ravindran.
\newblock Bridge correlational neural networks for multilingual multimodal
  representation learning.
\newblock In Kevin Knight, Ani Nenkova, and Owen Rambow, editors, \emph{{NAACL}
  {HLT} 2016, The 2016 Conference of the North American Chapter of the
  Association for Computational Linguistics: Human Language Technologies, San
  Diego California, USA, June 12-17, 2016}, pages 171--181. The Association for
  Computational Linguistics, 2016.
\newblock \doi{10.18653/v1/n16-1021}.
\newblock URL \url{https://doi.org/10.18653/v1/n16-1021}.

\bibitem[Ramnath and
  Hasegawa{-}Johnson(2020)]{DBLP:journals/corr/abs-2012-15484}
Kiran Ramnath and Mark Hasegawa{-}Johnson.
\newblock Seeing is knowing! fact-based visual question answering using
  knowledge graph embeddings.
\newblock \emph{CoRR}, abs/2012.15484, 2020.
\newblock URL \url{https://arxiv.org/abs/2012.15484}.

\bibitem[Ramnath et~al.(2021)Ramnath, Sari, Hasegawa-Johnson, and
  Yoo]{ramnath-etal-2021-worldly}
Kiran Ramnath, Leda Sari, Mark Hasegawa-Johnson, and Chang Yoo.
\newblock Worldly wise ({W}o{W}) - cross-lingual knowledge fusion for
  fact-based visual spoken-question answering.
\newblock In \emph{Proceedings of the 2021 Conference of the North American
  Chapter of the Association for Computational Linguistics: Human Language
  Technologies}, pages 1908--1919, Online, June 2021. Association for
  Computational Linguistics.
\newblock \doi{10.18653/v1/2021.naacl-main.153}.
\newblock URL \url{https://aclanthology.org/2021.naacl-main.153}.

\bibitem[Ramponi and Plank(2020)]{DBLP:conf/coling/RamponiP20}
Alan Ramponi and Barbara Plank.
\newblock Neural unsupervised domain adaptation in {NLP} - {A} survey.
\newblock In Donia Scott, N{\'{u}}ria Bel, and Chengqing Zong, editors,
  \emph{Proceedings of the 28th International Conference on Computational
  Linguistics, {COLING} 2020, Barcelona, Spain (Online), December 8-13, 2020},
  pages 6838--6855. International Committee on Computational Linguistics, 2020.
\newblock \doi{10.18653/v1/2020.coling-main.603}.
\newblock URL \url{https://doi.org/10.18653/v1/2020.coling-main.603}.

\bibitem[Reimers and Gurevych(2019)]{DBLP:conf/emnlp/ReimersG19}
Nils Reimers and Iryna Gurevych.
\newblock Sentence-bert: Sentence embeddings using siamese bert-networks.
\newblock In Kentaro Inui, Jing Jiang, Vincent Ng, and Xiaojun Wan, editors,
  \emph{Proceedings of the 2019 Conference on Empirical Methods in Natural
  Language Processing and the 9th International Joint Conference on Natural
  Language Processing, {EMNLP-IJCNLP} 2019, Hong Kong, China, November 3-7,
  2019}, pages 3980--3990. Association for Computational Linguistics, 2019.
\newblock \doi{10.18653/v1/D19-1410}.
\newblock URL \url{https://doi.org/10.18653/v1/D19-1410}.

\bibitem[Reiter and Dale(1997)]{DBLP:journals/nle/ReiterD97}
Ehud Reiter and Robert Dale.
\newblock Building applied natural language generation systems.
\newblock \emph{Nat. Lang. Eng.}, 3\penalty0 (1):\penalty0 57--87, 1997.
\newblock \doi{10.1017/S1351324997001502}.
\newblock URL \url{https://doi.org/10.1017/S1351324997001502}.

\bibitem[Rinsche(2005)]{rinsche-2005-computer}
Adriane Rinsche.
\newblock Computer-assisted multingual {E}-communication in a variety of
  application areas.
\newblock In \emph{Proceedings of Machine Translation Summit X: Posters}, pages
  458--464, Phuket, Thailand, September 13-15 2005.
\newblock URL \url{https://aclanthology.org/2005.mtsummit-posters.20}.

\bibitem[Rotman et~al.(2018)Rotman, Vuli{\'c}, and
  Reichart]{rotman-etal-2018-bridging}
Guy Rotman, Ivan Vuli{\'c}, and Roi Reichart.
\newblock Bridging languages through images with deep partial canonical
  correlation analysis.
\newblock In \emph{Proceedings of the 56th Annual Meeting of the Association
  for Computational Linguistics (Volume 1: Long Papers)}, pages 910--921,
  Melbourne, Australia, July 2018. Association for Computational Linguistics.
\newblock \doi{10.18653/v1/P18-1084}.
\newblock URL \url{https://aclanthology.org/P18-1084}.

\bibitem[Route et~al.(2019)Route, Hillis, Czeresnia~Etinger, Zhang, and
  Black]{route-etal-2019-multimodal}
James Route, Steven Hillis, Isak Czeresnia~Etinger, Han Zhang, and Alan~W
  Black.
\newblock Multimodal, multilingual grapheme-to-phoneme conversion for
  low-resource languages.
\newblock In \emph{Proceedings of the 2nd Workshop on Deep Learning Approaches
  for Low-Resource NLP (DeepLo 2019)}, pages 192--201, Hong Kong, China,
  November 2019. Association for Computational Linguistics.
\newblock \doi{10.18653/v1/D19-6121}.
\newblock URL \url{https://aclanthology.org/D19-6121}.

\bibitem[Rust et~al.(2021)Rust, Pfeiffer, Vulic, Ruder, and
  Gurevych]{DBLP:conf/acl/RustPVRG20}
Phillip Rust, Jonas Pfeiffer, Ivan Vulic, Sebastian Ruder, and Iryna Gurevych.
\newblock How good is your tokenizer? on the monolingual performance of
  multilingual language models.
\newblock In Chengqing Zong, Fei Xia, Wenjie Li, and Roberto Navigli, editors,
  \emph{Proceedings of the 59th Annual Meeting of the Association for
  Computational Linguistics and the 11th International Joint Conference on
  Natural Language Processing, {ACL/IJCNLP} 2021, (Volume 1: Long Papers),
  Virtual Event, August 1-6, 2021}, pages 3118--3135. Association for
  Computational Linguistics, 2021.
\newblock \doi{10.18653/v1/2021.acl-long.243}.
\newblock URL \url{https://doi.org/10.18653/v1/2021.acl-long.243}.

\bibitem[Sanabria et~al.(2018)Sanabria, Caglayan, Palaskar, Elliott, Barrault,
  Specia, and Metze]{DBLP:journals/corr/abs-1811-00347}
Ramon Sanabria, Ozan Caglayan, Shruti Palaskar, Desmond Elliott, Lo{\"{\i}}c
  Barrault, Lucia Specia, and Florian Metze.
\newblock How2: {A} large-scale dataset for multimodal language understanding.
\newblock \emph{CoRR}, abs/1811.00347, 2018.
\newblock URL \url{http://arxiv.org/abs/1811.00347}.

\bibitem[Shi et~al.(2019)Shi, Mao, Gimpel, and Livescu]{shi-etal-2019-visually}
Haoyue Shi, Jiayuan Mao, Kevin Gimpel, and Karen Livescu.
\newblock Visually grounded neural syntax acquisition.
\newblock In Anna Korhonen, David~R. Traum, and Llu{\'{\i}}s M{\`{a}}rquez,
  editors, \emph{Proceedings of the 57th Conference of the Association for
  Computational Linguistics, {ACL} 2019, Florence, Italy, July 28- August 2,
  2019, Volume 1: Long Papers}, pages 1842--1861. Association for Computational
  Linguistics, 2019.
\newblock \doi{10.18653/v1/p19-1180}.
\newblock URL \url{https://doi.org/10.18653/v1/p19-1180}.

\bibitem[Shimizu et~al.(2018)Shimizu, Rong, and
  Miyazaki]{shimizu-etal-2018-visual}
Nobuyuki Shimizu, Na~Rong, and Takashi Miyazaki.
\newblock Visual question answering dataset for bilingual image understanding:
  A study of cross-lingual transfer using attention maps.
\newblock In \emph{Proceedings of the 27th International Conference on
  Computational Linguistics}, pages 1918--1928, Santa Fe, New Mexico, USA,
  August 2018. Association for Computational Linguistics.
\newblock URL \url{https://aclanthology.org/C18-1163}.

\bibitem[Singh et~al.(2021)Singh, Sanayai~Meetei, Singh, and
  Bandyopadhyay]{singh-etal-2021-multiple}
Salam~Michael Singh, Loitongbam Sanayai~Meetei, Thoudam~Doren Singh, and Sivaji
  Bandyopadhyay.
\newblock Multiple captions embellished multilingual multi-modal neural machine
  translation.
\newblock In \emph{Proceedings of the First Workshop on Multimodal Machine
  Translation for Low Resource Languages (MMTLRL 2021)}, pages 2--11, Online
  (Virtual Mode), September 2021. INCOMA Ltd.
\newblock URL \url{https://aclanthology.org/2021.mmtlrl-1.2}.

\bibitem[Su et~al.(2020)Su, Zhu, Cao, Li, Lu, Wei, and
  Dai]{DBLP:conf/iclr/SuZCLLWD20}
Weijie Su, Xizhou Zhu, Yue Cao, Bin Li, Lewei Lu, Furu Wei, and Jifeng Dai.
\newblock {VL-BERT:} pre-training of generic visual-linguistic representations.
\newblock In \emph{8th International Conference on Learning Representations,
  {ICLR} 2020, Addis Ababa, Ethiopia, April 26-30, 2020}. OpenReview.net, 2020.
\newblock URL \url{https://openreview.net/forum?id=SygXPaEYvH}.

\bibitem[Sur{\'{\i}}s et~al.(2022)Sur{\'{\i}}s, Epstein, and
  Vondrick]{DBLP:conf/cvpr/SurisEV22}
D{\'{\i}}dac Sur{\'{\i}}s, Dave Epstein, and Carl Vondrick.
\newblock Globetrotter: Connecting languages by connecting images.
\newblock In \emph{{IEEE/CVF} Conference on Computer Vision and Pattern
  Recognition, {CVPR} 2022, New Orleans, LA, USA, June 18-24, 2022}, pages
  16453--16463. {IEEE}, 2022.
\newblock \doi{10.1109/CVPR52688.2022.01598}.
\newblock URL \url{https://doi.org/10.1109/CVPR52688.2022.01598}.

\bibitem[Susanto et~al.(2021)Susanto, Wang, Yadav, Jain, and
  Htun]{susanto-etal-2021-rakutens}
Raymond~Hendy Susanto, Dongzhe Wang, Sunil Yadav, Mausam Jain, and Ohnmar Htun.
\newblock Rakuten{'}s participation in {WAT} 2021: Examining the effectiveness
  of pre-trained models for multilingual and multimodal machine translation.
\newblock In \emph{Proceedings of the 8th Workshop on Asian Translation
  (WAT2021)}, pages 96--105, Online, August 2021. Association for Computational
  Linguistics.
\newblock \doi{10.18653/v1/2021.wat-1.9}.
\newblock URL \url{https://aclanthology.org/2021.wat-1.9}.

\bibitem[Trmal et~al.(2017)Trmal, Wiesner, Peddinti, Zhang, Ghahremani, Wang,
  Manohar, Xu, Povey, and Khudanpur]{DBLP:conf/interspeech/TrmalWPZGWMXPK17}
Jan Trmal, Matthew Wiesner, Vijayaditya Peddinti, Xiaohui Zhang, Pegah
  Ghahremani, Yiming Wang, Vimal Manohar, Hainan Xu, Daniel Povey, and Sanjeev
  Khudanpur.
\newblock The kaldi openkws system: Improving low resource keyword search.
\newblock In Francisco Lacerda, editor, \emph{Interspeech 2017, 18th Annual
  Conference of the International Speech Communication Association, Stockholm,
  Sweden, August 20-24, 2017}, pages 3597--3601. {ISCA}, 2017.
\newblock URL
  \url{http://www.isca-speech.org/archive/Interspeech\_2017/abstracts/0601.html}.

\bibitem[Tsai et~al.(2019)Tsai, Bai, Liang, Kolter, Morency, and
  Salakhutdinov]{DBLP:conf/acl/TsaiBLKMS19}
Yao{-}Hung~Hubert Tsai, Shaojie Bai, Paul~Pu Liang, J.~Zico Kolter,
  Louis{-}Philippe Morency, and Ruslan Salakhutdinov.
\newblock Multimodal transformer for unaligned multimodal language sequences.
\newblock In Anna Korhonen, David~R. Traum, and Llu{\'{\i}}s M{\`{a}}rquez,
  editors, \emph{Proceedings of the 57th Conference of the Association for
  Computational Linguistics, {ACL} 2019, Florence, Italy, July 28- August 2,
  2019, Volume 1: Long Papers}, pages 6558--6569. Association for Computational
  Linguistics, 2019.
\newblock \doi{10.18653/v1/p19-1656}.
\newblock URL \url{https://doi.org/10.18653/v1/p19-1656}.

\bibitem[Vendrov et~al.(2016)Vendrov, Kiros, Fidler, and
  Urtasun]{DBLP:journals/corr/VendrovKFU15}
Ivan Vendrov, Ryan Kiros, Sanja Fidler, and Raquel Urtasun.
\newblock Order-embeddings of images and language.
\newblock In Yoshua Bengio and Yann LeCun, editors, \emph{4th International
  Conference on Learning Representations, {ICLR} 2016, San Juan, Puerto Rico,
  May 2-4, 2016, Conference Track Proceedings}, 2016.
\newblock URL \url{http://arxiv.org/abs/1511.06361}.

\bibitem[Vilares et~al.(2020)Vilares, G{\'o}mez-Rodr{\'\i}guez,
  Fern{\'a}ndez-N{\'u}{\~n}ez, Penas, and Viteri]{vilares-etal-2020-bringing}
Jes{\'u}s Vilares, Carlos G{\'o}mez-Rodr{\'\i}guez, Lu{\'\i}s
  Fern{\'a}ndez-N{\'u}{\~n}ez, Dar{\'\i}o Penas, and Jorge Viteri.
\newblock Bringing roguelikes to visually-impaired players by using {NLP}.
\newblock In \emph{Workshop on Games and Natural Language Processing}, pages
  59--67, Marseille, France, May 2020. European Language Resources Association.
\newblock ISBN 979-10-95546-40-5.
\newblock URL \url{https://aclanthology.org/2020.gamnlp-1.9}.

\bibitem[Wang et~al.(2019{\natexlab{a}})Wang, Huang, Celikyilmaz, Gao, Shen,
  Wang, Wang, and Zhang]{DBLP:conf/cvpr/WangHcGSWWZ19}
Xin Wang, Qiuyuan Huang, Asli Celikyilmaz, Jianfeng Gao, Dinghan Shen,
  Yuan{-}Fang Wang, William~Yang Wang, and Lei Zhang.
\newblock Reinforced cross-modal matching and self-supervised imitation
  learning for vision-language navigation.
\newblock In \emph{{IEEE} Conference on Computer Vision and Pattern
  Recognition, {CVPR} 2019, Long Beach, CA, USA, June 16-20, 2019}, pages
  6629--6638. Computer Vision Foundation / {IEEE}, 2019{\natexlab{a}}.
\newblock \doi{10.1109/CVPR.2019.00679}.
\newblock URL
  \url{http://openaccess.thecvf.com/content\_CVPR\_2019/html/Wang\_Reinforced\_Cross-Modal\_Matching\_and\_Self-Supervised\_Imitation\_Learning\_for\_Vision-Language\_Navigation\_CVPR\_2019\_paper.html}.

\bibitem[Wang et~al.(2019{\natexlab{b}})Wang, Wu, Chen, Li, Wang, and
  Wang]{DBLP:conf/iccv/WangWCLWW19}
Xin Wang, Jiawei Wu, Junkun Chen, Lei Li, Yuan{-}Fang Wang, and William~Yang
  Wang.
\newblock Vatex: {A} large-scale, high-quality multilingual dataset for
  video-and-language research.
\newblock In \emph{2019 {IEEE/CVF} International Conference on Computer Vision,
  {ICCV} 2019, Seoul, Korea (South), October 27 - November 2, 2019}, pages
  4580--4590. {IEEE}, 2019{\natexlab{b}}.
\newblock \doi{10.1109/ICCV.2019.00468}.
\newblock URL \url{https://doi.org/10.1109/ICCV.2019.00468}.

\bibitem[Weiss et~al.(2017)Weiss, Chorowski, Jaitly, Wu, and
  Chen]{DBLP:conf/interspeech/WeissCJWC17}
Ron~J. Weiss, Jan Chorowski, Navdeep Jaitly, Yonghui Wu, and Zhifeng Chen.
\newblock Sequence-to-sequence models can directly translate foreign speech.
\newblock In Francisco Lacerda, editor, \emph{Interspeech 2017, 18th Annual
  Conference of the International Speech Communication Association, Stockholm,
  Sweden, August 20-24, 2017}, pages 2625--2629. {ISCA}, 2017.
\newblock URL
  \url{http://www.isca-speech.org/archive/Interspeech\_2017/abstracts/0503.html}.

\bibitem[Willemsen et~al.(2018)Willemsen, de~Wit, Krahmer, de~Haas, and
  Vogt]{willemsen-etal-2018-context}
Bram Willemsen, Jan de~Wit, Emiel Krahmer, Mirjam de~Haas, and Paul Vogt.
\newblock Context-sensitive natural language generation for robot-assisted
  second language tutoring.
\newblock In \emph{Proceedings of the Workshop on {NLG} for Human{--}Robot
  Interaction}, pages 1--7, Tilburg, The Netherlands, November 2018.
  Association for Computational Linguistics.
\newblock \doi{10.18653/v1/W18-6901}.
\newblock URL \url{https://aclanthology.org/W18-6901}.

\bibitem[Xu et~al.(2020{\natexlab{a}})Xu, Cao, Wang, Chen, Zhou, Zeng, Wang,
  Chen, Yin, Zhang, Jiang, Wang, and Li]{xu-etal-2020-xiaomingbot}
Runxin Xu, Jun Cao, Mingxuan Wang, Jiaze Chen, Hao Zhou, Ying Zeng, Yuping
  Wang, Li~Chen, Xiang Yin, Xijin Zhang, Songcheng Jiang, Yuxuan Wang, and Lei
  Li.
\newblock {X}iaomingbot: {A} {M}ultilingual {R}obot {N}ews {R}eporter.
\newblock In \emph{Proceedings of the 58th Annual Meeting of the Association
  for Computational Linguistics: System Demonstrations}, pages 1--8, Online,
  July 2020{\natexlab{a}}. Association for Computational Linguistics.
\newblock \doi{10.18653/v1/2020.acl-demos.1}.
\newblock URL \url{https://aclanthology.org/2020.acl-demos.1}.

\bibitem[Xu et~al.(2020{\natexlab{b}})Xu, Zhu, Shi, Zeng, and
  Huang]{DBLP:conf/ijcnlp/XuZSZH20}
Ruochen Xu, Chenguang Zhu, Yu~Shi, Michael Zeng, and Xuedong Huang.
\newblock Mixed-lingual pre-training for cross-lingual summarization.
\newblock In Kam{-}Fai Wong, Kevin Knight, and Hua Wu, editors,
  \emph{Proceedings of the 1st Conference of the Asia-Pacific Chapter of the
  Association for Computational Linguistics and the 10th International Joint
  Conference on Natural Language Processing, {AACL/IJCNLP} 2020, Suzhou, China,
  December 4-7, 2020}, pages 536--541. Association for Computational
  Linguistics, 2020{\natexlab{b}}.
\newblock URL \url{https://aclanthology.org/2020.aacl-main.53/}.

\bibitem[Yoshikawa et~al.(2017)Yoshikawa, Shigeto, and
  Takeuchi]{DBLP:conf/acl/YoshikawaST17}
Yuya Yoshikawa, Yutaro Shigeto, and Akikazu Takeuchi.
\newblock {STAIR} captions: Constructing a large-scale japanese image caption
  dataset.
\newblock In Regina Barzilay and Min{-}Yen Kan, editors, \emph{Proceedings of
  the 55th Annual Meeting of the Association for Computational Linguistics,
  {ACL} 2017, Vancouver, Canada, July 30 - August 4, Volume 2: Short Papers},
  pages 417--421. Association for Computational Linguistics, 2017.
\newblock \doi{10.18653/v1/P17-2066}.
\newblock URL \url{https://doi.org/10.18653/v1/P17-2066}.

\bibitem[Zarrie{\ss} et~al.(2016)Zarrie{\ss}, Hough, Kennington,
  Manuvinakurike, DeVault, Fern{\'a}ndez, and
  Schlangen]{zarriess-etal-2016-pentoref}
Sina Zarrie{\ss}, Julian Hough, Casey Kennington, Ramesh Manuvinakurike, David
  DeVault, Raquel Fern{\'a}ndez, and David Schlangen.
\newblock {P}ento{R}ef: A corpus of spoken references in task-oriented
  dialogues.
\newblock In \emph{Proceedings of the Tenth International Conference on
  Language Resources and Evaluation ({LREC}'16)}, pages 125--131,
  Portoro{\v{z}}, Slovenia, May 2016. European Language Resources Association
  (ELRA).
\newblock URL \url{https://aclanthology.org/L16-1019}.

\bibitem[Zhang et~al.(2017)Zhang, Dai, Tuytelaars, Moens, and
  Gool]{DBLP:journals/corr/ZhangDTMG17}
Ted Zhang, Dengxin Dai, Tinne Tuytelaars, Marie{-}Francine Moens, and Luc~Van
  Gool.
\newblock Speech-based visual question answering.
\newblock \emph{CoRR}, abs/1705.00464, 2017.
\newblock URL \url{http://arxiv.org/abs/1705.00464}.

\bibitem[Zhang et~al.(2020)Zhang, Chen, Wang, Utiyama, Sumita, Li, and
  Zhao]{DBLP:conf/iclr/0001C0USLZ20}
Zhuosheng Zhang, Kehai Chen, Rui Wang, Masao Utiyama, Eiichiro Sumita, Zuchao
  Li, and Hai Zhao.
\newblock Neural machine translation with universal visual representation.
\newblock In \emph{8th International Conference on Learning Representations,
  {ICLR} 2020, Addis Ababa, Ethiopia, April 26-30, 2020}. OpenReview.net, 2020.
\newblock URL \url{https://openreview.net/forum?id=Byl8hhNYPS}.

\bibitem[Zhou et~al.(2018)Zhou, Cheng, Lee, and Yu]{zhou-etal-2018-visual}
Mingyang Zhou, Runxiang Cheng, Yong~Jae Lee, and Zhou Yu.
\newblock A visual attention grounding neural model for multimodal machine
  translation.
\newblock In \emph{Proceedings of the 2018 Conference on Empirical Methods in
  Natural Language Processing}, pages 3643--3653, Brussels, Belgium,
  October-November 2018. Association for Computational Linguistics.
\newblock \doi{10.18653/v1/D18-1400}.
\newblock URL \url{https://aclanthology.org/D18-1400}.

\bibitem[Zhou et~al.(2021)Zhou, Zhou, Wang, Cheng, Li, Yu, and
  Liu]{DBLP:conf/cvpr/ZhouZW0LYL21}
Mingyang Zhou, Luowei Zhou, Shuohang Wang, Yu~Cheng, Linjie Li, Zhou Yu, and
  Jingjing Liu.
\newblock {UC2:} universal cross-lingual cross-modal vision-and-language
  pre-training.
\newblock In \emph{{IEEE} Conference on Computer Vision and Pattern
  Recognition, {CVPR} 2021, virtual, June 19-25, 2021}, pages 4155--4165.
  Computer Vision Foundation / {IEEE}, 2021.
\newblock \doi{10.1109/CVPR46437.2021.00414}.
\newblock URL
  \url{https://openaccess.thecvf.com/content/CVPR2021/html/Zhou\_UC2\_Universal\_Cross-Lingual\_Cross-Modal\_Vision-and-Language\_Pre-Training\_CVPR\_2021\_paper.html}.

\bibitem[Zhu et~al.(2019)Zhu, Wang, Wang, Zhou, Zhang, Wang, and
  Zong]{DBLP:conf/emnlp/ZhuWWZZWZ19}
Junnan Zhu, Qian Wang, Yining Wang, Yu~Zhou, Jiajun Zhang, Shaonan Wang, and
  Chengqing Zong.
\newblock {NCLS:} neural cross-lingual summarization.
\newblock In Kentaro Inui, Jing Jiang, Vincent Ng, and Xiaojun Wan, editors,
  \emph{Proceedings of the 2019 Conference on Empirical Methods in Natural
  Language Processing and the 9th International Joint Conference on Natural
  Language Processing, {EMNLP-IJCNLP} 2019, Hong Kong, China, November 3-7,
  2019}, pages 3052--3062. Association for Computational Linguistics, 2019.
\newblock \doi{10.18653/v1/D19-1302}.
\newblock URL \url{https://doi.org/10.18653/v1/D19-1302}.

\end{thebibliography}

\end{document}